\RequirePackage{fix-cm}

\documentclass[twocolumn]{svjour3}          

\smartqed  

\usepackage{amsmath}
\usepackage{amssymb}
\usepackage{graphicx}

\usepackage{arydshln}
\usepackage{url}
\usepackage{color}

\usepackage{url}

\def\Eq#1{Equation (\ref{eqn:#1})}
\def\Fig#1{Figure \ref{fig:#1}}
\def\Table#1{Table \ref{table:#1}}
\def\Sec#1{Section \ref{sec:#1}}
\begin{document}

\title{Train Sparsely, Generate Densely: Memory-efficient Unsupervised Training of High-resolution Temporal GAN}

\author{Masaki Saito \and Shunta Saito \and Masanori Koyama \and Sosuke Kobayashi}

\institute{Masaki Saito \at
              \email{msaito@preferred.jp}
           \and
           Shunta Saito \at
              \email{shunta@preferred.jp}
           \and
           Masanori Koyama \at
              \email{masomatics@preferred.jp}
           \and
           Sosuke Kobayashi \at
              \email{sosk@preferred.jp}
}

\date{}
\journalname{IJCV}

\maketitle

\begin{abstract}
Training of Generative Adversarial Network (GAN) on a video dataset is a challenge because of the sheer size of the dataset and the complexity of each observation.
In general, the computational cost of training GAN scales exponentially with the resolution.
In this study, we present a novel memory efficient method of unsupervised learning of high-resolution video dataset whose computational cost scales only linearly with the resolution.
We achieve this by designing the generator model as a stack of small sub-generators and training the model in a specific way.
We train each sub-generator with its own specific discriminator.
At the time of the training, we introduce between each pair of consecutive sub-generators an auxiliary subsampling layer that reduces the frame-rate by a certain ratio.
This procedure can allow each sub-generator to learn the distribution of the video at different levels of resolution.
We also need only a few GPUs to train a highly complex generator that far outperforms the predecessor in terms of inception scores.
The source code is available at \url{https://github.com/pfnet-research/tgan2}.
\end{abstract}

\section{Introduction}
\label{sec:introduction}
Generative Adversarial Network (GAN) is a powerful family of unsupervised learning, and various versions of GANs have been developed to date for different types of datasets, including image and audio dataset.
GANs have been particularly successful for its application to image dataset \cite{Radford2016,Miyato2018,Brock2018}.

In this study, we present a novel method of unsupervised learning for video dataset, an important type of dataset with numerous applications such as autonomous vehicles, creative tasks, video compression, and frame interpolation.

There are two major challenges in training a generative model for video dataset.
The first challenge comes from the sheer complexity of each observation.
Video has a time dimension in addition to width and height, and the correlation between each pair of time frames are usually governed by complex dynamics underlying the system.
Also, many applications of video generation methods--including those pertaining to industrial projects--require every frame of the generated video to be photo-realistic.
This is a challenging problem on its own because photo-realistic image generation has been made possible only recently with the invention of techniques to stabilize the training of GANs on large dataset \cite{Karras2018,Miyato2018,Mescheder2018}.
One must not only prepare a model that is sophisticated enough to produce a photo-realistic video but also to come up with an appropriate strategy to train the model stably within a reasonable time frame and computational resource.

The second challenge is the size of the dataset to be used in the training process.
Unsupervised learning of generative model usually requires large dataset to guarantee high generalization ability.
This can be a particular problem for the learners of video dataset because each observation is \textit{large} on its own.
Moreover, the required computational resource grows exponentially with the resolution.
A naive approach is bound to fail because one must reserve massive resource for both \textit{high-resolution dataset} and the set of \textit{model parameters}.

Most previous studies have only focused on improving the photo-realism of the output \cite{Vondrick2016,Saito2017,Tulyakov2018} while using the dataset of low-resolution videos in the range of $64 \times 64$ for the training.

In general, batch-size is limited by the memory capacity of GPU, and under a situation where the number of usable GPU is limited, it is unrealistic in practice to use a sufficiently large batch size for the training of currently available models on $256 \times 256$ resolution videos with few GPUs.
These challenges warrant a clever mechanism to train a massive model fast in a memory-efficient manner.

\begin{figure}[t]
\centering
  \includegraphics[width=1.0\linewidth]{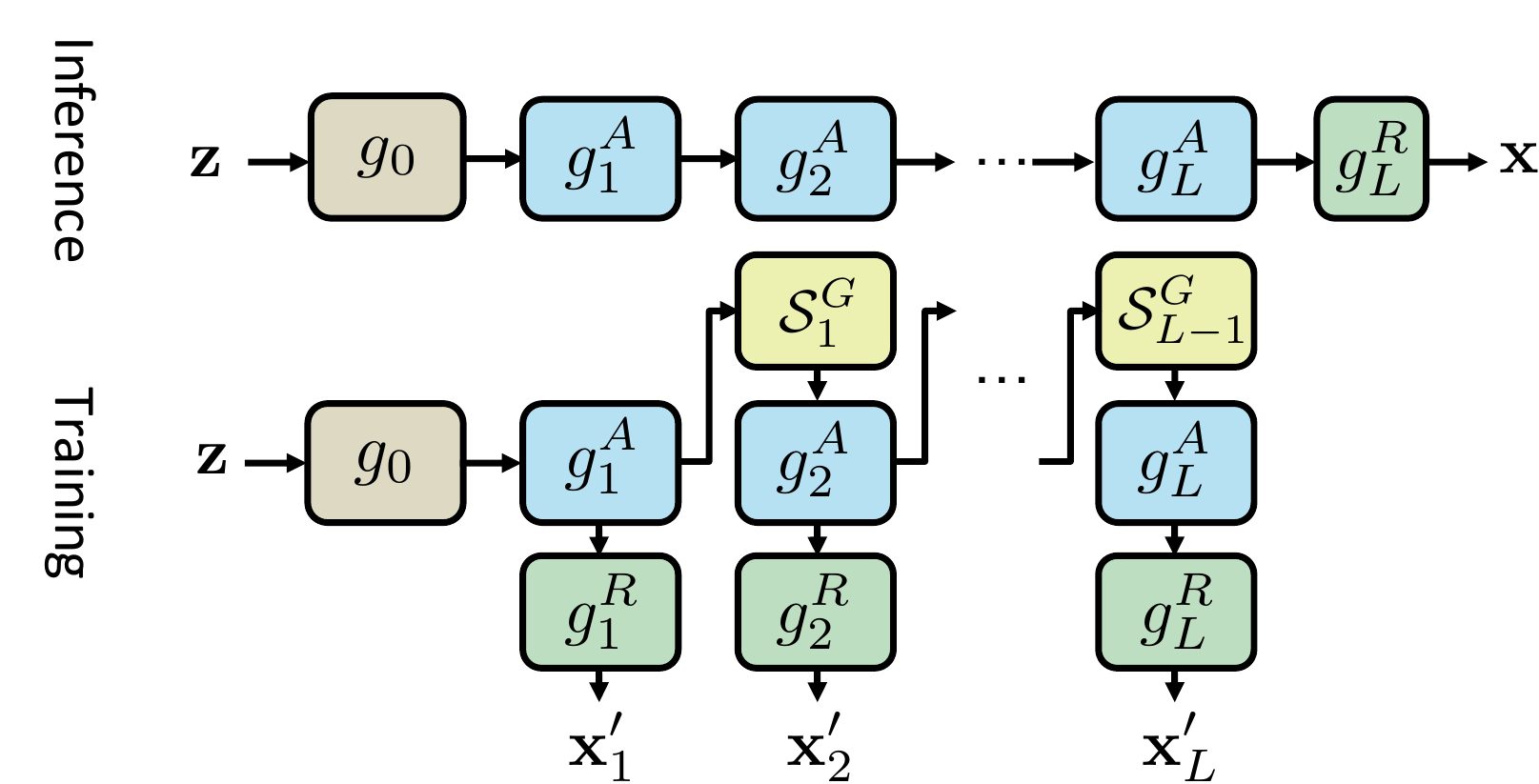}
  \caption{The generator using multiple subsampling layers.}
  \label{fig:multigen}

\end{figure}

We resolve these two problems by introducing a separate architecture into the model at the time of the training.
Our generator is a stack of multiple sub-generators, in which each sub-generator takes a video of intermediate feature maps as an input and outputs a video of more complex intermediate features with the same number of frames (\Fig{multigen}).
At the training process, we introduce a subsampling layer between each sub-generator and force the intermediate features from each sub-generator to make a detour to a subsampling layer that reduces its frame-rate by a certain ratio.
For example, if the video of intermediate features at $k-1$th layer has $m$ frames, the sub-generator at $k$-th level will receive a sparse, subsampled version of the video with $m / s_t$ frames at the time of the training, where $s_t \geq 1$ denotes the subsampling rate.
This way, the sub-generators at low level are made to receive high-frame-rate, low-resolution videos of abstract and global information, and the sub-generators at a high level are made to receive low-frame-rate, high-resolution videos of local information.
This procedure can significantly reduce the computational cost and the memory requirement during the training process.
Also, by preparing a separate discriminator for each sub-generator, we evaluate the fidelity of the generated video at various levels of resolution.

Because local information tends to be slow in changing (low frame-rate), our design of the {\em division of roles} allows us to produce high-resolution videos with high fidelity while keeping the memory consumption low.
Our method can efficiently train a generator that can generate videos with significantly higher inception score ($\sim 26$) than all predecessors.

\section{Related work}
\label{sec:related_work}

\subsection{Image generation}

Many applications of the Generative Adversarial Network \cite{Goodfellow2014} mainly focus on the image generation problem; this is primarily because that the Deep Convolutional Generative Adversarial Networks (DCGAN) \cite{Radford2016} demonstrated that the GAN is quite effective in image generation.
After that, the DCGAN was extended to a network called Self-Attention Generative Adversarial Networks (SAGAN) \cite{Zhang2018} that includes spectral normalization \cite{Miyato2018} and self-attention blocks \cite{Wang2018b}.
BigGANs \cite{Brock2018} succeeded in generating high fidelity images by introducing several tricks such as orthogonal regularization that stabilizes the training with large batch size.
In the image-to-image translation problem, which transfers an input image from its domain to another domain,
there exist many studies for transforming a high-resolution image to another high-resolution image. \cite{Isola2017,Wang2018a,Zhu2017,Liu2017,Huang2018}.

\subsection{Video-to-video translation}

Recently, several studies have succeeded in converting high-resolution videos into other ones in a different domain. RecycleGAN \cite{Bansal2018} extends a concept of CycleGAN \cite{Zhu2017}, and solves a video retargeting problem with two video datasets for which correspondences are not given.
Vid2Vid \cite{Wang2018} learns a model that maps a source domain into an output one from a pair of videos and generates high-resolution videos.
Contrary to these models that can be trained well with a small batch size (e.g., one or two),
image and video generation GAN requires a large batch size for the training, which makes the training of GAN models for video generation more difficult.

\subsection{Multi-scale GANs}

Our approach is related to methods in which a generator produces multi-scale images.
LAPGAN \cite{Denton2015} first generates a coarse image and updates it by using a difference of an initial image.
StackGAN \cite{Zhang2017,Zhang2017a} and HDGAN \cite{Zhang2018a} directly generates multi-scale images, and the corresponding discriminator returns a score from each image.
Although our approach itself is similar to StackGAN and HDGAN,
the most significant difference lies in the existence of sampling layers;
it is useful for problems where the number of dimensions in a sample is quite large.

ProgressiveGAN \cite{Karras2018} is another model that uses multi-resolution images
and generates high-resolution images by growing both the generator and the discriminator gradually.
Although the ProgressiveGAN also saves computational cost by using low-resolution images in the early stages of the training, our method has two advantages compared to ProgressiveGAN.
The first advantage is that our method does not require hyperparameters to grow the network according to the number of iterations dynamically. It also means that it does not need to dynamically change the batch size according to the number of iterations.
The second is that our method can generate a large sample from the beginning regardless of the number of iterations. It is useful for evaluating inception score \cite{Salimans2016} using a pre-trained model accepting only a sample with a fixed shape.

\subsection{Video prediction}

Video prediction, which estimates subsequent images from a given few frames, is one of the major problems in computer vision. Although there is no unified view on how to model the domain of video, many studies dealing with video prediction problem directly predict the next frame with recurrent networks such as LSTM
\cite{Ranzato2014,Oh2015,Srivastava2015,Kalchbrenner2016,Finn2016,Lotter2017,Ebert2017,Babaeizadeh2018,Byeon2018,Denton2018,Lee2018}.
Another well-known approach is to predict intermediate features of videos such as optical flow \cite{Liang2017,Liu2017a,Hao2018,Li2018}.
Some studies introduce adversarial training to avoid generating a blurred image caused by the image reconstruction loss \cite{Mathieu2016,Liang2017,Lee2018}.

\subsection{Video generation}

There are two major approaches in the field of video generation using GANs.
One is a study for generating videos by incorporating dataset specific prior knowledge into the method (e.g., limiting the problem to human action generation and giving the number of parts as prior knowledge) \cite{Cai2018,Zhao2018,Yang2018},
and the other is a study that can handle any datasets without such restriction \cite{Vondrick2016,Saito2017,Ohnishi2018,Tulyakov2018,Acharya2018}.
Our study is related to the latter.

We describe several models for video generation in the following.
To the best of our knowledge, VGAN \cite{Vondrick2016} is the first model to generate videos using GANs and consists of a 2D network to generate a still background and 3D convolutional networks to generate foreground videos.
After that, Saito et al. \cite{Saito2017} found that it is better to separate a spatiotemporal generator into time-series and space models to generate videos and proposed TGAN that first generates a set of latent vectors corresponding to each frame and then transforms them into actual images.
MoCoGAN \cite{Tulyakov2018} was proposed to produce videos more efficiently by decomposing the latent space into the motion and the content subspaces.
Unlike TGAN where the discriminator consists of a stack of 3D convolutional layers, MoCoGAN uses two different sub-discriminators to improve the quality of videos; the first is a three-dimensional discriminator that aims to extract global motion in the video, and another is two-dimensional one that identifies from a still frame in the video.

Although these models illustrate that models using GAN are also useful in video generation, they have a problem of requiring an enormous computational cost and GPU memory. It is particularly problematic when dealing with high-resolution video. Our proposed model aims to solve it.
Besides, our model can be regarded as a further extension of the MoCoGAN, where the discriminator uses two different discriminators.
Although our model also uses multiple discriminators, its computational cost and memory consumption are quite lower than MoCoGAN, which directly takes the generated video as an input. It is due to the multiple subsampling layers described later.

\section{Method}

In this section, we first describe the conventional design of GANs for video dataset.
We will describe the challenges posed by these models, and explain how our models can resolve them.

\subsection{GAN models for video generation}
Generative Adversarial Network is a framework of unsupervised learning in which a generator model strives to mimic the target distribution while a discriminator model strives to distinguish the samples from the target distribution from the samples synthesized by the generator.
The generator synthesizes an artificial sample by applying a network function $G$ to a sample from an user-specified prior distribution $p_z$.
The discriminator network $D$ classifies an input as \textit{real} or \textit{synthetic}.
Let us denote a sample from $p_z$ by $ \vec{z}$, and denote a sample from $p_d$ (target distribution) by $\vec{x}$.
The training of these networks is performed by
alternately maximizing and minimizing the following objective:
\begin{align}
\label{eqn:GAN}
    \mathbb{E}_{\vec{x} \sim p_d}[\ln D(\vec{x})] +
    \mathbb{E}_{\vec{z} \sim p_z}[\ln (1 - D(G(\vec{z})))],
\end{align}

\begin{figure}[t]
\centering
  \includegraphics[width=0.8\linewidth]{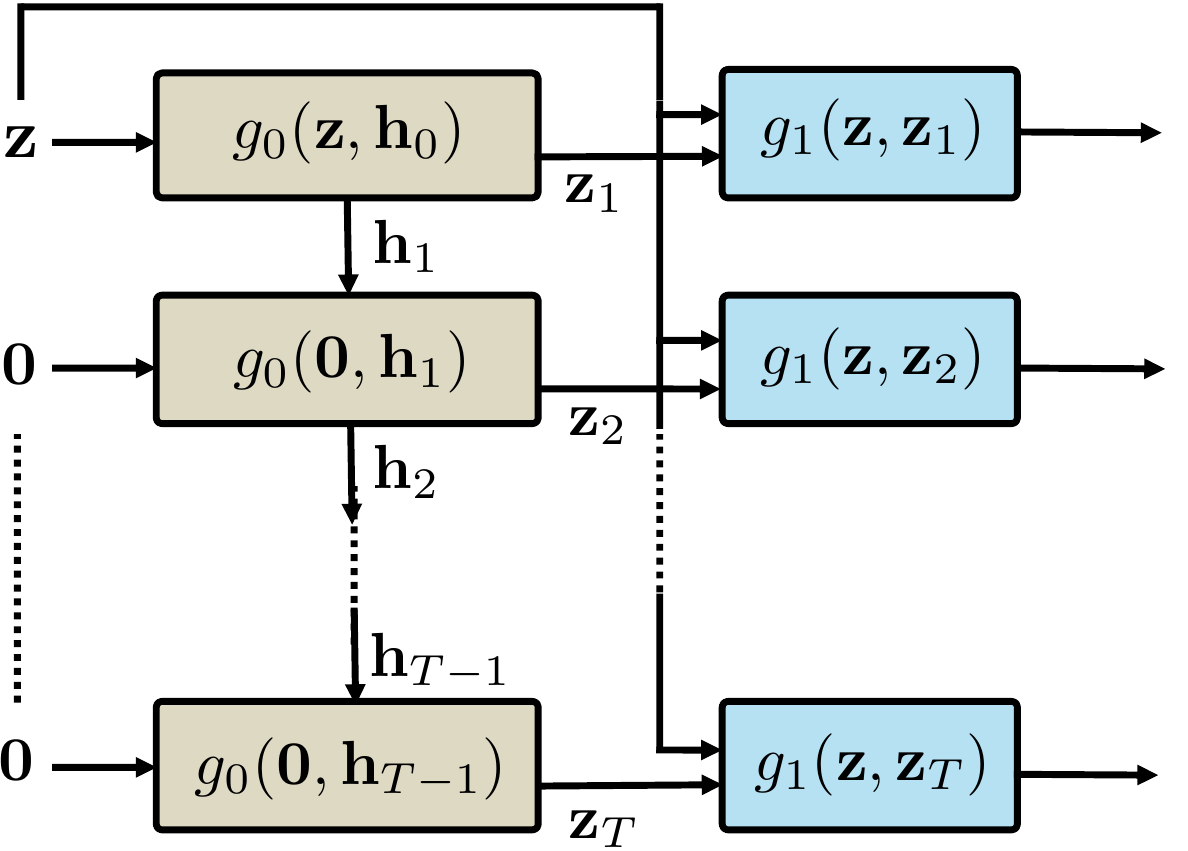}
  \caption{The overview of relationship between a temporal generator and an image generator used in conventional temporal GANs. To produce a set of $T$ latent vectors, the temporal generator typically is a recurrent network. For the proposed method, we used a convolutional LSTM for this part as detailed in \Sec{generator_architecture}.}
  \label{fig:temporal_generator}

\end{figure}

Conventional temporal GAN (i.e., TGAN and MoCoGAN) uses a generator consisting of two sub-networks: {\em temporal generator} $g_0$ and {\em image generator} $g_1$.
For the generation of a $T$-frame video, {\em temporal generator} generates a size-$T$ set of latent vectors (or a $T$-frame video in the latent space)
$\{\vec{z}_1, \dots, \vec{z}_T\}$ from noise vector $\vec{z}$, and {\em image generator} transforms the \textit{video} of latent vectors and the noise vector into the \textit{video} of images as shown in \Fig{temporal_generator}.
The synthesized video is then classified as \textit{synthetic} or \textit{real} by a discriminator network consisting of three-dimensional convolutional layers.

\subsection{Subsampling layer}

As mentioned in the introduction, the conventional architecture introduced above is memory inefficient.
A memory that must be reserved for the three dimensional convolutional layer in the discriminator is especially heavy, and the number of parameters in the model can be literally on order 100 million.
Because GPU must reserve a large space for massive $G$ and $D$, currently available lines of GPU cannot process a sufficient batch when a video is larger than $64 \times 64$ pixels with 16 frames.

In order to resolve this resource shortage, we introduce an architecture called \textit{subsampling layer}s at the time of training.

For simplicity, we first describe the training of GAN with a single subsampling layer.
Suppose a generator that outputs video $\vec{x}$ from noise vector $\vec{z}$ and consists of two blocks: {\em abstract block} $g^A$ and {\em rendering block} $g^R$.
The abstract block computes the latent feature map (we call it an {\it abstract map}) from noise vector, and the rendering block transforms it into a video.
In inference time, the generation process of samples by this generator is equivalent to that of
the generator in the conventional GAN; that is, $G(\vec{z})$ can be represented by
\begin{align}
    \vec{x} = G(\vec{z}) = \left( g^R \circ g^A \right) (\vec{z}).
\end{align}

The discriminator in conventional temporal GANs computes a score from a high-frame rate video of original resolution.
This can be especially costly when the original resolution is high.
To cope with this problem, in the training stage, we modify $G$ into $G'$ as follows by introducing between $g^R$ and $g^A$ a {\it subsampling layer} ${\cal S}^G$
that reduces the frame rate of the abstract map produced from $g^A$.

With $G'$, the subsampled video $\vec{x}'$ is produced by applying $g^R$ to  ${\cal S}^G(g^A(\vec{z}))$. That is,
\begin{align}
    \vec{x}' &= G'(\vec{z}) = \left( g^R \circ {\cal S}^G \circ g^A \right) (\vec{z}).
\end{align}
The discriminator $D'$ then evaluates the score for $\vec{x}'$.

In the training process of $D$, one must also prepare a set of \textit{real} data to compare against the synthetic data.
We prepare this by applying $S^D$ to the real videos, an architecture that downscales the resolution and reduces the frame-rate so that the dimension of the output tensor matches the output of $G'$.

In other words, in the training stage of our method, the objective of \Eq{GAN}
can be rewritten by
\begin{align}
    \mathbb{E}_{\vec{x} \sim p_d}[\ln D'({\cal S}^D(\vec{x}))] +
    \mathbb{E}_{\vec{z} \sim p_z}[\ln (1 - D'(G'(\vec{z})))].
\end{align}
This way, we can significantly reduce the burden on the discriminator.
The computational cost for $D'$ is much smaller than the original $D$.

Indeed, this benefit comes at the cost of making the domain of the video at the inference-time differ from that of the video at the training time.
We introduce two tricks to deal with this problem.
The first trick is to statically change the position of the first frame (that is, we sample video at times $t s_t + b_t$ with random $b_t$; cf. \Eq{subsampling2}) so that there will be no frames that are not sampled.
The second trick is to use multiple subsampling layers that drop the frame rate by different scales.
We describe the detail of this method in the next subsection.

\subsection{Multiple subsampling layers}
\label{sec:multi_abstract_layers}

Unfortunately, however, we experimentally observed that the scores of the samples generated by the architecture in the previous section are significantly lower than those generated by the naive implementation without the subsampling layer.

By using a model made of multiple sub-generators and multiple subsampling layers, however, we can drastically reduce the overall computational cost without compromising the quality of the video.
In fact, the quality of the video generated by our stacked sub-generator is significantly better than that of the video generated by the conventional architecture.
Below we elaborate the construction of the model (\Fig{multigen}).

We describe the model consisting of $L$ sub-generators, or $L$ abstract blocks.
At the time of the training, this architecture is trained with $L$ corresponding rendering blocks and $L-1$ subsampling layers.

Let us denote the abstract block, rendering block, and the subsampling layer at level $l$ by $g^A_l$, $g^R_l$, and ${\cal S}^G_l$ respectively.
In the inference time, $\vec{x}$ can be evaluated simply by sequentially applying abstract blocks and rendering with a single $g^R_L$, i.e.,
\begin{align}
    \vec{x} = \left( g^R_L \circ g^A_L \circ g^A_{L - 1} \circ \cdots \circ g^A_1 \right) (\vec{z}).
\end{align}
Here and also in the following explanation, $g^A_1$ includes the temporal generator $g_0$, but we abuse the notation $g^A_1$ as $g^A_1 \circ g_0$ for simplicity (see \Fig{multigen}).
Note that the noise vector is concatenated with each latent vector from the temporal generator only when it is input to the first image generator $g^A_1$.
In the training time, we use $L$-set of sub-generators consisting of $G'_l(\vec{z})$, each of which recursively applies $g^A_m$ and ${\cal S}^G_m$ ($m=1, \dots, l - 1$) to abstract maps and converting the output of the final abstract block $g^A_l$ to the video by $g^R_l$:
\begin{align}
\label{eqn:multigen}
    G'_1 &= g_1^R \circ g_1^A \nonumber \\
    G'_2 &= g_2^R \circ g_2^A \circ \left( {\cal S}^G_1 \circ g_1^A \right) \nonumber \\
    &\;\;\vdots \nonumber \\
    G'_L &= g_L^R \circ g_L^A \circ \left( {\cal S}^G_{L-1} \circ g_{L-1}^A \right) \circ \cdots \circ \left( {\cal S}^G_1 \circ g_1^A \right).
\end{align}
Note that in our implementation, all rendering blocks ($g_i^R (i=1,\dots,L)$) do not have any shared parameters, i.e., every block has its own parameters.
$G'_1$ is a model that generates a video of lowest resolution and highest-frame-rate.
It applies $g_1^A$ to the high-frame-rate video of latent vectors and applies $g_1^R$ to the result to produce a high-frame-rate video of low-resolution images $\vec{x}'_1$.

$G'_2$ on the other hand reduces the frame-rate of the video output of $g_1^A$ with ${\cal S}_1^G$, feed the subsampled video to $g_2^A$, and converts the output of $g_2^A$ to a video of higher resolution image $\vec{x}'_2$ by $g_2^R$.
Defined recursively, $G'_L$ generates a video of highest resolution and lowest-frame-rate.
Because the increase in the resolution is countered by the decrease in frame-rate, the computational cost for $G_L$ does not increase exponentially with the resolution of the video.

\subsection{Multiple discriminators}

In order to train the above generator consisting of multiple sub-generators $G'_1, \dots, G'_L$, we need a discriminator that evaluates a score for the set of videos produced by them.
Our discriminator consists of multiple sub-discriminators.
Let $D'_l$ be the $l$-th sub-discriminator that takes a sample $x_l'$  from $l$-th sub-generator $G'_l$ and returns a scalar value.
Our discriminator $D'$ evaluates a score for a set of $x_l'$s by the following formula:
\vspace{-1truemm}
\begin{align}
    D'(\vec{x}'_1, \dots, \vec{x}'_L) &= \sigma \left( \sum_{l=1}^L D'_l(\vec{x}'_l) \right),
\end{align}
where $\sigma(\cdot)$ is a sigmoid function.
To take a sample from the raw dataset and evaluate a score for it, we apply ${\cal S}^D_l$ to each raw video, which downscales the resolution and reduces the frame-rate, so that the output will match the synthetic video produced by $G_l'$.

\subsection{Role of each discriminator}

The essence of our method is the division of roles;
instead of feeding the dense raw dataset of a single massive model,
we train each sub-generator with select parts of the dataset that can help the sub-generator improve at playing the given role.

The role of the sub-generator with low indices is to mimic the original video dataset at an abstract level; that is, to produce a low-resolution video that flows naturally with time.
The discriminators with low indices are responsible for evaluating the quality of high-frame-rate videos of low resolution.
This allows the generators with low indices to capture global motion in the video that is independent from high resolution details.

On the other hand, the role of the sub-generator with high indices is to mimic the original video dataset in visual quality.
That is, they only require a low frame-rate video of high resolution to train.
The discriminator with the highest index in our model may in fact be designed to evaluate the fidelity of the still image (one frame).

Our model can also be regarded as an extension of MoCoGAN, a model that uses two different discriminators.

Our method also aims to reduce both computational cost and memory consumption simultaneously.
We will further discuss this topic in \Sec{cost_and_memory}.

\section{Network architecture}
\label{sec:videonet}
\begin{figure*}[t!]
\centering
  \includegraphics[width=1.0\linewidth]{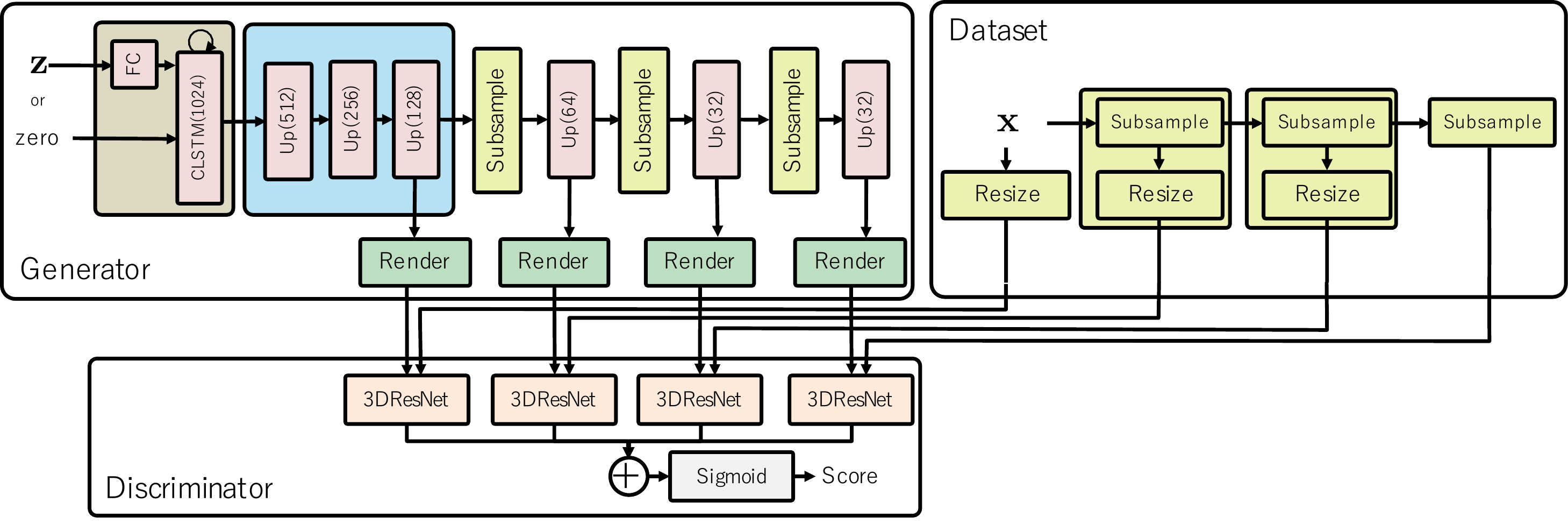}
  \caption{Network configuration of our model. ``CLSTM($C$)'' represents the convolutional LSTM with $C$ channels and $3 \times 3$ kernel. ``Up($C$)'' means the upsampling block that returns a feature map with $C$ channels and twice the resolution of the input.}
  \label{fig:network}

\end{figure*}
\begin{figure*}[t]
\centering
  \includegraphics[width=0.95\linewidth]{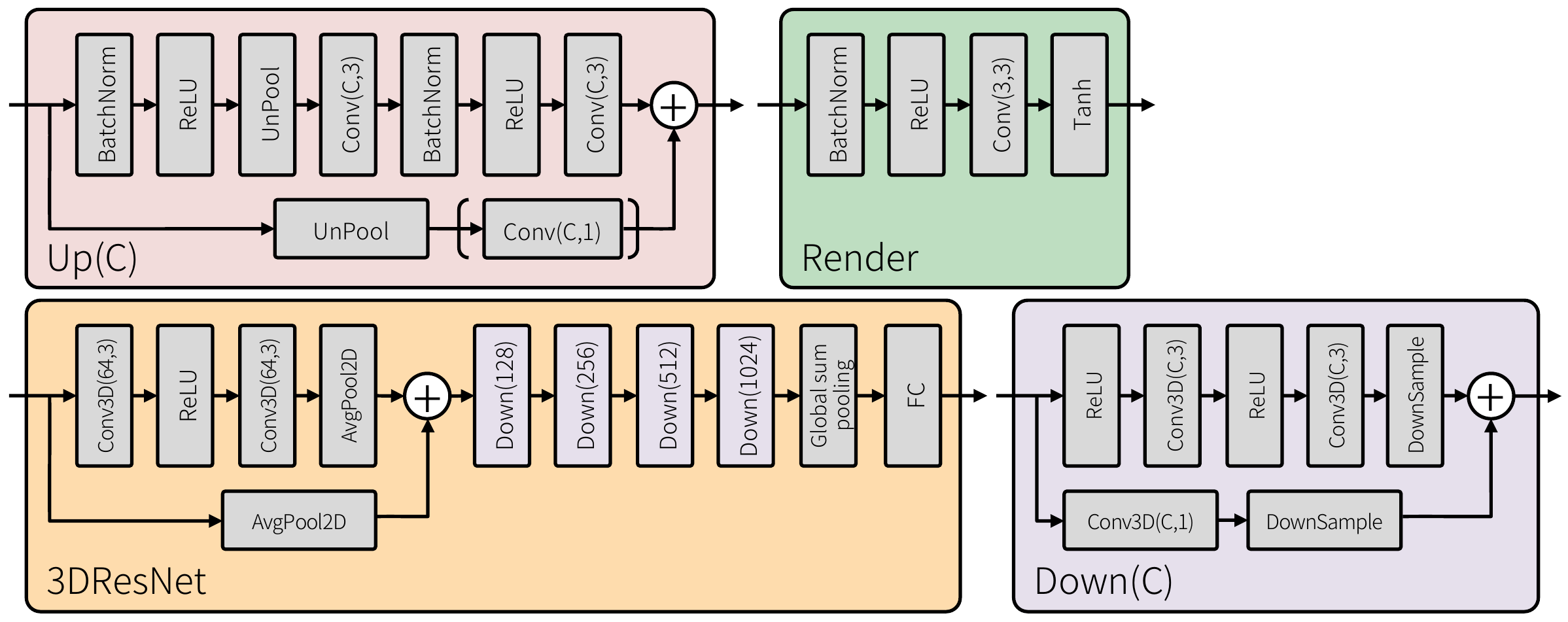}
  \caption{Details of the blocks used in the main paper. ``Conv($C, k$)'' denotes a 2D convolutional layer with $C$ channels and $(k \times k)$ kernel. ``Conv3D($C, k$)'' denotes a 3D convolutional layer with $C$ channels and $(k \times k \times k)$ kernel. ``UnPool'' denotes a 2D unpooling layer with $(2 \times 2)$ kernel and stride 2. ``AvgPool2D'' denotes a 2D  average pooling layer along the spatial dimensions with $(2 \times 2)$ kernel and stride 2. Note that it does not perform pooling operation along the temporal dimension. ``DownSample'' means the downsampling operator. If the size of each dimension of the input 3D feature maps is larger than one, this operator performs the average pooling along its axis (if the size is odd when performing the average pooling, the padding of the target axis is set to one). Otherwise, average pooling is not performed for that axis. $(\cdot)$ in ``Up$(C)$'' means that the blocks in the bracket are not inserted if the number of input channels is equivalent to that of output channels.}
  \label{fig:blocks}
\end{figure*}

\subsection{Generator}
\label{sec:generator_architecture}
We describe our design of a generator consisting of four sub-generators that synthesizes a video with $T$ frames and $W \times H$ pixels.
As with the TGAN \cite{Saito2017} and MoCoGAN \cite{Tulyakov2018},
our generator first produces $T$ latent feature maps from a noise vector and then transforms each map into a corresponding frame (see \Fig{temporal_generator}).
The specific network structure is shown in \Fig{network} and \Fig{blocks}.

We first describe the flow of the inference. Given $d$-dimensional noise vector
$\vec{z}$ randomly drawn from the uniform distribution within a range of $[-1, 1]$,
the generator converts it into a feature map of $(W / 64) \times (H / 64)$ pixel resolution through a fully-connected layer.
A recurrent block consisting of a Convolutional LSTM \cite{Shi2015} (CLSTM) receives this feature map as an input,
and then returns another feature map with the same shape.
Note that at $t=0$ the CLSTM receives the feature map derived from $\vec{z}$ as input,
but at $t \geq 1$ it always receives a feature map derived from a zero vector
(i.e., $\vec{z}$ is used to initialize the state of the CLSTM).
After that, each feature map is transformed into another feature map with $W \times H$ pixels by six upsampling blocks, and a rendering block renders it into the frame
(As we mentioned before, all $g_i^R (i=1,\dots,L)$ do not share any parameters and have their own parameters).
That is, the upsampling block is a function that outputs a feature map whose resolution is double that of the input feature map, and the rendering block converts it to the image while maintaining the resolution.

In the training phase, the generator progressively reduces the size of abstract maps
with the three subsampling layers placed between each $g^A_i$.

Each layer is given the role of reducing the number of frames.

Consider for example an abstract map $\vec{h}$ with $C_h$ channels,
$T_h$ frames, width $W_h$ and height $H_h$.
We represent each element of this tensor by $h_{c,t,h,w}$.
The subsampling layer with reduction rate $s_t$ reduces the shape of $\vec{h}$ to
$(C_h \times \lceil T_h / s_t \rceil \times H_h \times W_h)$; 
The output $\vec{h}'$ produced from subsampling layer is the following probabilistic mixture of subsampled versions of $h_{c,t,h,w}$:
\begin{align}
\label{eqn:subsampling}
    h'_{c, \tau, h, w} = \sum_{t} B(\tau, n, t) h_{c, t, h, w},
\end{align}
where $B$ is a Boolean function given by
\begin{align}
\label{eqn:subsampling2}
  B(\tau, n, t) = \begin{cases}
    1 & \mathrm{if}\;\;\tau = t s_t + b_t\\
    0 & \mathrm{otherwise}
  \end{cases}
\end{align}
with $b_t$ being a sample from discrete uniform distribution  ${\cal U}\{0, \min(s_t, T_h) - 1\}$.
Using the slicing notation of NumPy \cite{Oliphant2015}, the above equation can be evaluated easily by invoking \texttt{hd = h[:, bt::st]}.
In experiments, we set $s_t$ to the same value for all subsampling layers.

The advantage of this strategy is that it can prevent the memory consumption of the discriminator from growing exponentially with the resolution.

For this example model, the amount of GPU memory required by the discriminator will be constant regardless of the resolution if $s_t \geq 4$ and $T$ is sufficiently large.

At the same time, if $s_t$ is too large relative to $T$, the number of frames will reduce to $1$ at a block close to the input layer, and there will be no computational gain from the subsampling layer from that point onward.
The number of original frames required to guarantee the \textit{constant-order computational cost} at every block grows exponentially with $s_t$.
It is therefore important to choose the appropriate pair of $s_t$ and the number of blocks that suits the purpose and the memory capacity of GPU.

In \Sec{experiments} we will investigate the effect of the choice of $s_t$ on the quality of the generated video.

\subsection{Discriminator}
\label{sec:discriminator}
As we described in \Sec{multi_abstract_layers}, our discriminator consists of
several sub-discriminators. In our implementation, we used four 3D ResNet models, each containing several spatiotemporal three-dimensional residual blocks \cite{Hara2018,He2016} and one fully-connected layer.
The network configuration of each 3D ResNet was almost the same as the discriminator used in Miyato et al. \cite{Miyato2018} except that the kernel of all convolutional layers was replaced with $(3 \times 3)$ to $(3 \times 3 \times 3)$. For more details, see \Fig{blocks}.
As for the initializers, we used the GlorotUniform initializer \cite{Glorot2010} with a scale $\sqrt{2}$ in the residual paths of both the ``Up($C$)'' and the ``Down($C$)'', and the normal GlorotUniform initializer in the shortcut paths.

We shall emphasize that, even though all sub\discretionary{-}{-}{-}
discriminators are given the same network structure, they all play different roles in the model as a whole.
The sub-discriminator with the lowest index evaluates the fidelity of the global flow of a given video, and the sub-discriminator with the highest index evaluates the photorealism of randomly selected several frames.

\subsection{Computational cost and Memory}
\label{sec:cost_and_memory}
\begin{figure*}
\begin{tabular}{cc}
    \!\!\!\!\includegraphics[width=0.5\linewidth]{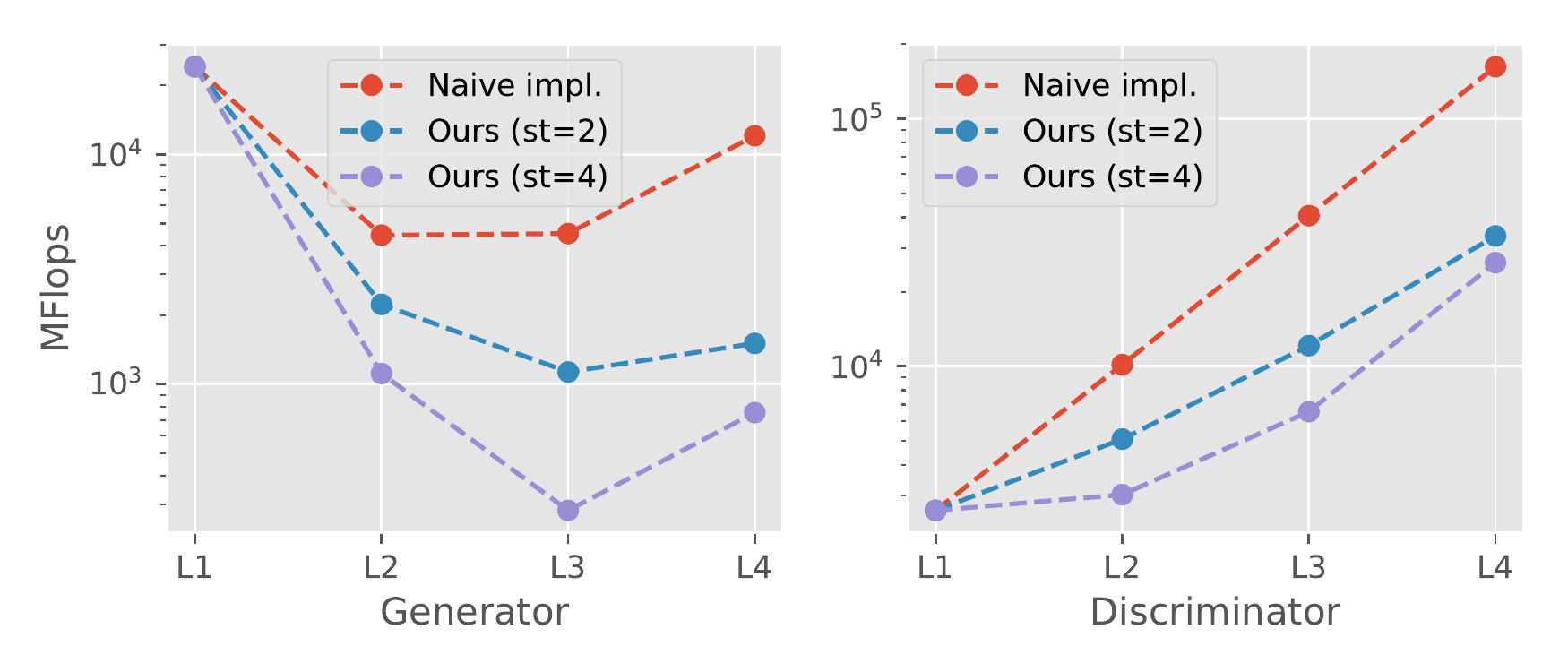} &
    \!\!\!\!\includegraphics[width=0.5\linewidth]{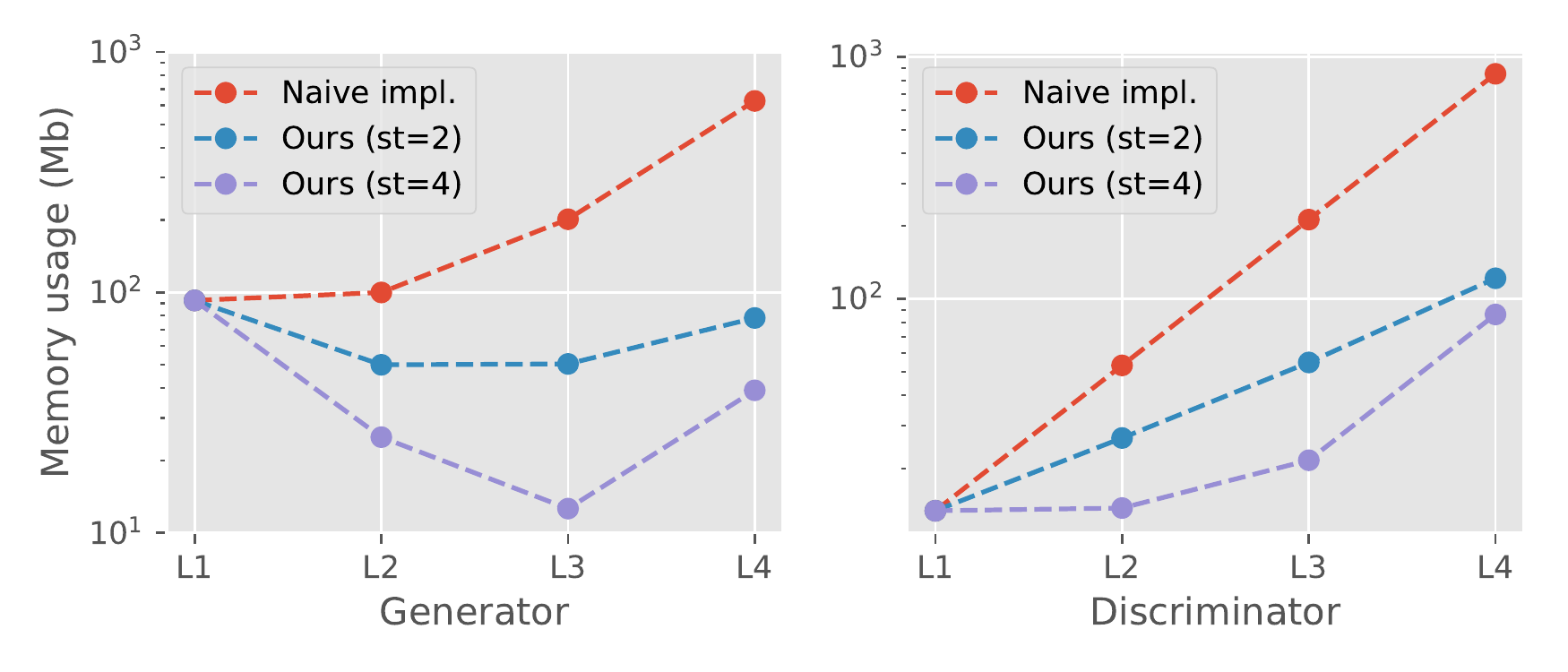} \\
    \!\!\!\!\text{\small (a) Computational cost of forward pass} &
    \!\!\!\!\text{\small (b) Total size of intermediate memory} \\
\end{tabular}
\caption{
Graphs representing the cost savings by enabling multiple subsampling layers.
The left figures show the theoretical computational cost of each block of the network, and the right figures denote the amount of intermediate memory in each block consumed by generated samples and hidden variables.
The horizontal axis of all graphs represents the level of the block.
Each block of the generator used to compute the cost contains layers inside the blue area in \Fig{network} and the corresponding rendering block.
Each block of the discriminator corresponds to each sub discriminator.
The number of frames used for generation is 16, and the batch size is 1.}
\label{fig:capacity}

\end{figure*}
\begin{table}
\centering
{\renewcommand{\arraystretch}{1.2}
\begin{tabular}{r|cc|cc}
Method & GFlops & Ratio & Memory & Ratio \\ \hline \hline
Gen (naive impl.) & 45.07 & & 1019 & \\
Subsampling ($s_t=2$) & 28.91 & 1.56x & 271 & 3.76x \\
Subsampling ($s_t=4$) & 26.20 & 1.72x & 169 & 6.03x \\ \hline
Dis (naive impl.) & 215.74 & & 1130 & \\
Subsampling ($s_t=2$) & 53.41 & 4.04x & 217 & 5.22x \\

Subsampling ($s_t=4$) & 38.40 & 5.62x & 135 & 8.37x \\ \hline
3D discriminator & 162.38 & & 851 & \\
3D + 2D dis. & 168.20 & & 921 & \\ \hline
\end{tabular}
}
\caption{The total computational cost and intermediate memory consumption in megabytes when enabling frame and batch reductions. The number of frames used for generation is 16, and the batch size is 1.}
\label{table:performance}
\end{table}

The computational cost and the memory consumption of the generator differ in nature from those of the discriminator (see \Fig{capacity} and \Table{performance}.)

While the computational cost of the generator is
almost constant for each block, the cost of the discriminator
increases exponentially with the level of the block.
On the other hand, the memory consumption of both generator and discriminator grows exponentially with the level.

This is problematic because the computational cost and the memory consumption of 3D ResNet are particularly high for the discriminator.

As shown in \Table{performance}, the computational cost for the discriminator tends to be significantly heavier than that of the generator
even when using a single 3D discriminator in the setting of ordinary GAN.

This is also true when using a discriminator similar to the one used in MoCoGAN
(``3D + 2D dis'' in \Table{performance}. See \Sec{baselines} for details.)
The computational cost for the discriminator grows exponentially with the dimension of the video,
and some countermeasure must be taken if one wishes to conduct unsupervised learning of high-resolution video with limited number of GPUs.

By successively reducing the frame-rate of the video with the aforementioned subsampling layer,
however, we can prevent the computational cost from growing exponentially.

When the number of frames reduced by the subsampling layer is
half of the original (i.e., $s_t=2$), we can improve the total memory usage
and computational cost of the model four$\sim$five times over the vanilla model without subsampling layers.

Meanwhile, this mechanism does not improve the computational cost of the generator as much.
However, most part of the overall computational cost comes from the discriminator,
and this mechanism alone can reduce the overall computational cost to a reasonable level.

In general, the computational gain increases with the size of $s_t$.
When the original number of frames is sufficiently large and $s_t \geq 4$, the computational cost for each sub-generator will be roughly kept constant because at each block, the number of frame decreases by the rate of $4$ while the height and the width doubles.

The computational gain is not as high in our implementation because the number of original frames used in our experiments was just $16$.
For example, \Fig{capacity} shows that the actual memory consumption at the level 4 in the generator is larger than that of the level 3 because the number of frames handled by the level 3 is one since the number of original frames is $16$.

Even still, the memory consumption of our model with $s_t=4$ was about $60\%$ better than that of $s_t=2$.

Thus, the computational gain from subsampling layer tends to increase with the size of $s_t$.

\subsection{Dataset}
We use three subsampling functions to make four subsampled videos from one example video in the dataset. For the sample to be used for the sub-discriminator with index$=1$, we only apply one resize function to lower the resolution of the video by one eighth.
Meanwhile, the last sub-discriminator receives a video that was produced by applying three subsampling layers defined in \Eq{subsampling} to the input example.
This transformation reduces the number of frames to $1 / s_t^3$ while maintaining the resolution.

\subsection{Regularizer}
To improve the performance of the generator, we augmented the loss of the discriminator at level $l$ with a zero-centered gradient penalty evaluated over the target data distribution \cite{Mescheder2018}, given by

\begin{align}
\label{eqn:gradient_penalty}
    R_1 = \lambda \sum_{l=1}^L \sum_{i=1}^{n_l} \Vert \nabla D'_l (\vec{x}'_l) \Vert^2,
\end{align}
where $\lambda$ is a weight and $n_l$ represents the batch size at level $l$.

\subsection{Conditional model}
\label{sec:cgan}

In the previous discussion, we only focus on the problem
where the generator can only accept a noise vector as an input.
However, as with the field of the image generation (c.f., BigGAN \cite{Brock2018}),
we can extend the proposed method to accept not only the noise vector but also a discrete integer label
representing a category of the dataset.
Giving such information as an argument of the generator and the discriminator,
we can further improve the quality of the generated videos.

Specifically, we extend the generator and the discriminator according to the following.
In the generator, we first transform the discrete label into one-hot vector $\vec{l}$,
concatenate it with noise vector $\vec{z}$ to obtain another vector represented by $[\vec{l}, \vec{z}]$.
Using this vector as the input of of the FC layer in \Fig{network},
the discrete label information can be propagated to the ConvLSTM.
Regarding the extension of the image generator (i.e., the upsampling blocks after the ConvLSTM),
we adopted the technique performed by Miyato et al. \cite{Miyato2018}; that is, we replace
all the batch normalization layers with conditional batch normalization layers to ensure that
all the blocks receive discrete label information.
In the discriminator, we add a perturbation vector for the label to the final layer of each sub-discriminator
according to a method of projection discriminator \cite{Miyato2018a}.

\section{Experiments}
\label{sec:experiments}

\subsection{Datasets}
We used the following two datasets in the experiments.

\paragraph{UCF101}
UCF101 is a common video dataset that contains
13,320 videos with $320 \times 240$ pixels and 101 different sport categories such as
{\it Baseball Pitch} \cite{Soomro2012}.
In the experiments, we randomly extracted 16 frames from the training dataset,
cropped a rectangle with $240 \times 240$ pixels from the center, resized it to $192 \times 192$ pixels,
and used it for training.
The values of all the samples are normalized to $[-1, 1]$.
To amplify the samples we randomly flipped video during the training.

\paragraph{FaceForensics}
Following to Wang et al. \cite{Wang2018}, we created the facial videos from FaceForensics \cite{Rossler2018}
containing 854 news videos with different reporters. Specifically, we first identified the position of the face with a mask video in the dataset, cropped only the area of the face, and resized it to $256 \times 256$ pixels. In training, we randomly extracted 16 frames from them and sent these frames to the discriminator.
As with UCF101, all the values are normalized to $[-1, 1]$.

\subsection{Hyperparameters}
We used the Adam \cite{Kingma2015} optimizer with the learning rate of $1.0 \times 10^{-4}$, decayed linearly to 0 over 100K iterations. We also employed $\beta_1 = 0.0$ and $\beta_2 = 0.9$ in the Adam optimizer.
The local batch size of each GPU was selected so it fills the GPU memory of NVIDIA Tesla P100 (12Gb).
The total batch size used for experiments depends on the experiments.
The number of updates of the discriminator for each iteration was set to one.
The number of dimensions of $\vec{z}$ was set to 256, and $\lambda$ in \Eq{gradient_penalty} was 0.5.
We implemented all models using Chainer \cite{Tokui2015} and ChainerMN \cite{Akiba2017}.
The total number of parameters in our model is about $2.0 \times 10^8$, which is larger than the number of parameters used in BigGAN ($1.6 \times 10^8$) \cite{Brock2018}.

We manually confirmed that the training time is approximately proportional to GFlops shown in \Table{performance} under the same number of GPUs.
The whole training time including snapshot and computation of the inception score was about 61 hours under the environment of four GPUs, batch size 32, and $s_t=2$.
The training for each model used in the experiments ended within two to four days.

\subsection{Baseline models}
\label{sec:baselines}

To confirm the effectiveness of our model, we introduced two baseline models that generate high-resolution videos.
One is a high-resolution model consisting of a single generator and a single discriminator.
Specifically, the network architecture of the generator in this model is almost identical to that of the proposed model,
but it does not contain three rendering blocks that output low-resolution videos (that is, the generator only outputs a single high-resolution video during training).
The network of the discriminator is the same as the sub-discriminator used in our method.
Since the number of parameters of the generator is almost the same as that of our model,

we can see the difference of performance between the simple high-resolution model and our multi-scale model.

The other is a baseline model where the discriminator uses two different sub-discriminators similar to MoCoGAN.
Specifically, although the network of the generator is the same as that of the model mentioned above,
this discriminator consists of the 2D sub\discretionary{-}{-}{-} discriminator in addition to the above 3D ones.
The configuration of the 2D sub-discriminator is equivalent to the model of Miyato et al. \cite{Miyato2018} used for image generation.

The 3D sub-discriminator discriminates real videos from generated videos by directly classifying videos,
while the 2D sub-discriminator only discriminates a randomly selected frame in videos.
Using this baseline model, we can see the performance of simply inserting
the subsampling layer only in the final layer instead of a middle layer.

To improve the quality of baseline models,
we performed a grid search to find the optimal $\lambda$ in \Eq{gradient_penalty}.
We finally set $\lambda = 10.0$ for all experiments using these baseline models.

\subsection{Qualitative comparison of generated videos}
\subsubsection{FaceForensics}
\label{sec:face_forensics}

To confirm the quality of the video generated by our model,
we trained our models and two baselines using the two datasets described above.
First, we trained these models using the FaceForensics \cite{Rossler2018} dataset.
In order to check the effect of $s_t$ in the subsampling layer,
we trained the model with two values $s_t = 2, 4$.
In the proposed model, the generator has four sub-generators and each of which generates videos at different resolution and different number of frames depending on the setting of subsampling layers (the value of $s_t$).
Here, we denote the series of numbers of frames of those videos from different sub-generators as a tuple of the number of frames such as [16, 8, 4, 2].
Therefore, in the case of $s_t=2$, the number of video frames produced by the generator is $[16, 8, 4, 2]$.
On the other hand, in the case of $s_t = 4$, the number of video frames is $[16, 4, 1, 1]$ according to the definition of \Eq{subsampling2}.
The batch size was set to 32 for both. Eight GPUs were used in this experiment.

\begin{figure*}
\begin{tabular}{cc}
    \!\!\!\!\includegraphics[width=0.5\linewidth]{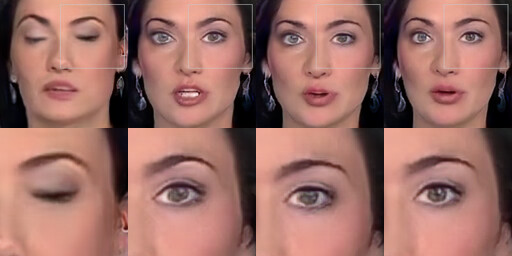} &
    \!\!\!\!\includegraphics[width=0.5\linewidth]{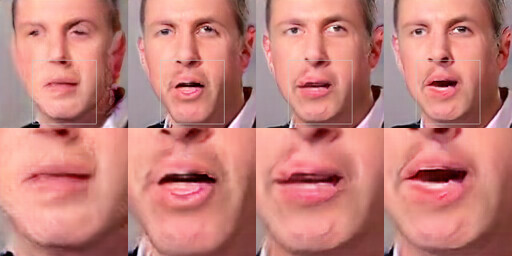} \\
    \!\!\!\!\includegraphics[width=0.5\linewidth]{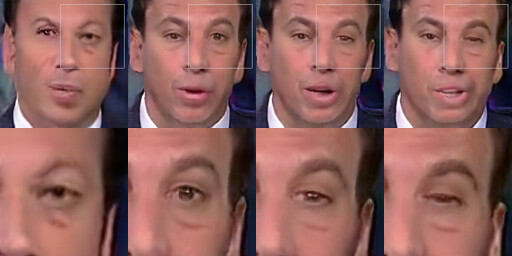} &
    \!\!\!\!\includegraphics[width=0.5\linewidth]{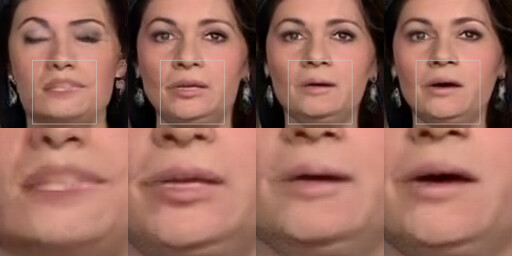} \\
    \!\!\!\!\text{\small Frame 1 \hspace{60mm} Frame 16} &
    \!\!\!\!\text{\small Frame 1 \hspace{60mm} Frame 16} \\
\end{tabular}
\caption{
An example of videos generated by our model ($s_t=2$) trained with FaceForensics dataset.
Every four frames out of 16 frames is shown in a row for the ease of identifing the motion in the video.
The top row represents the frames of the whole video.
The bottom row shows a magnified view of the area of the white box in the top row.
}
\label{fig:face_forensics}

\end{figure*}

The videos generated by our model are shown in \Fig{face_forensics}.
We confirmed that at $s_t=2$, our model generated high-resolution videos without collapsing the parts in a face.
Specifically, we observed that our model was able to generate high-fidelity images as a single still image, and facial parts such as eyes and mouth move smoothly.
To show clearly that our generated video is not just a sequence that continues a single still image, we also placed an enlarged view of the eyes and mouth in \Fig{face_forensics}.

\begin{figure*}
\begin{tabular}{c}
    \!\!\!\!\includegraphics[width=1.0\linewidth]{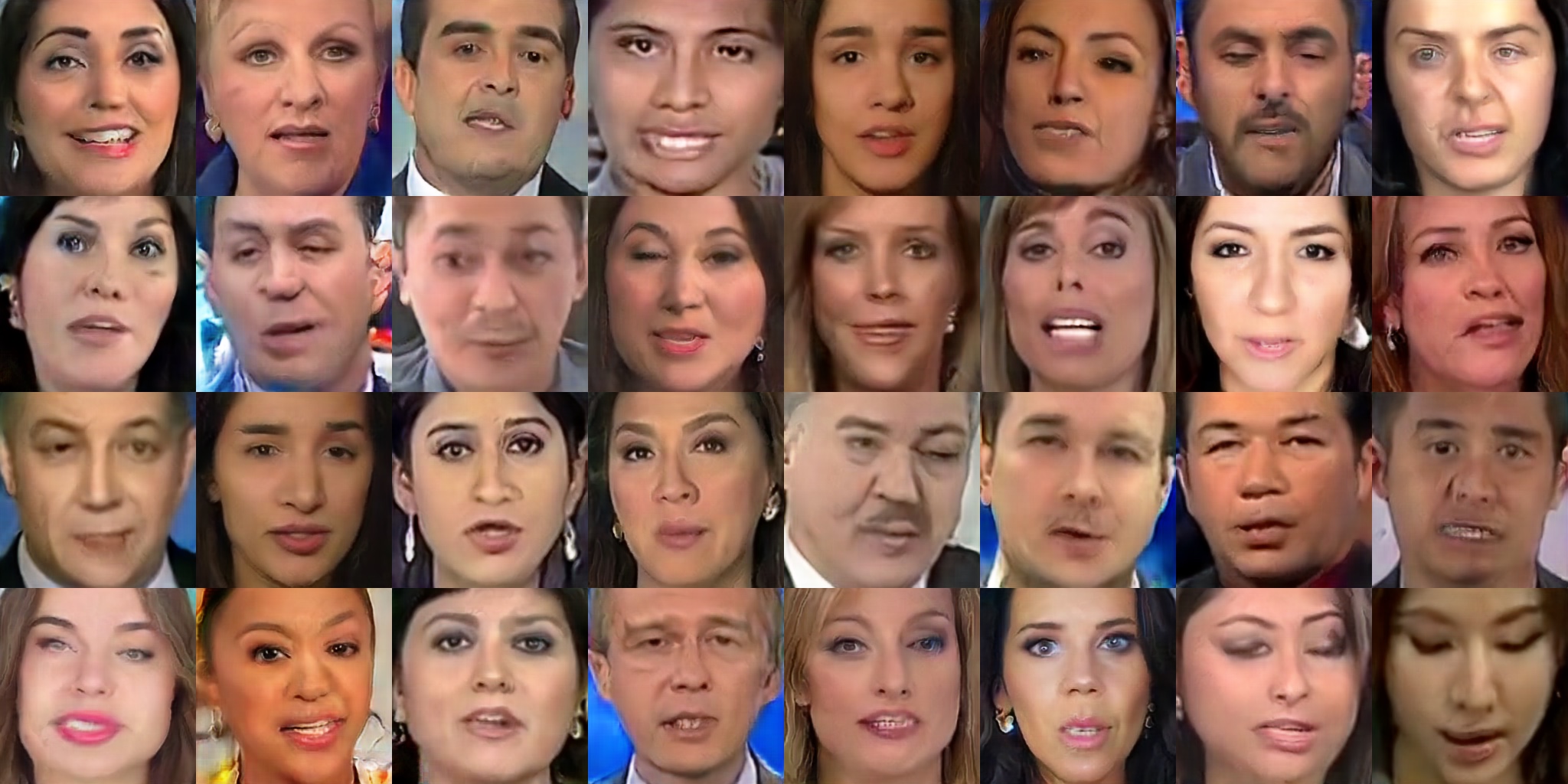} \\
\end{tabular}
\caption{
A list of still images extracted from videos generated by our model ($s_t=2$).
The FaceForensics dataset was used for the training.
}
\label{fig:face_forensics_1frames}

\end{figure*}

In addition to this, we also confirmed that the proposed method could generate
videos of various faces without causing mode collapse.
To show this diversity, we show several samples generated by our model in \Fig{face_forensics_1frames}.

As for the facial video, MoCoGAN also shows similar qualitative results using a similar facial video dataset, but its resolution is relatively small ($64 \times 64$ pixels).
We confirmed that even though our model handles samples that consume 16 times more memory than the conventional small samples, it can be efficiently trained while saving memory and computational cost with the subsampling layer.
As mentioned in Sections \ref{sec:introduction} and \ref{sec:cost_and_memory},
if one tries to use MoCoGAN that does not exploit the subsampling layer to generate
high resolution videos, there is a problem that the amount of consumed GPU memory is enormous (for example, MoCoGAN requires about 10Gb of memory to generate videos with $64 \times 64$ px even if batch size is 8).
The result of \Fig{face_forensics} shows that our model can be properly trained even at such a high resolution without extremely increasing the amount of memory consumed by the discriminator.

\begin{figure}
\begin{tabular}{cc}
    \!\!\!\!\rotatebox{90}{\parbox{3.0cm}{Our model ($s_t = 2$)}} &
    \!\!\!\!\includegraphics[width=0.9\linewidth]{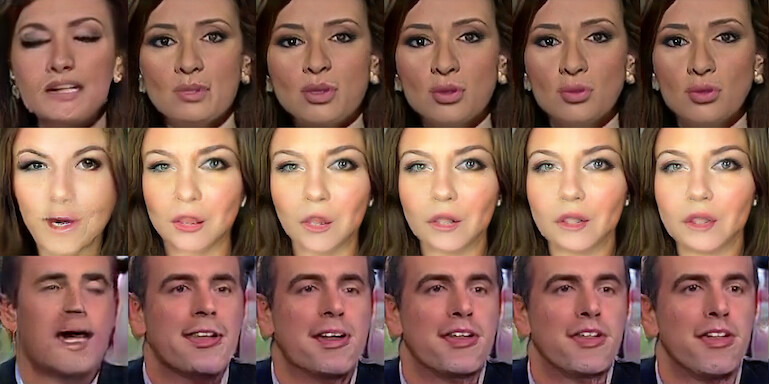}\\
    \!\!\!\!\rotatebox{90}{\parbox{3.0cm}{Our model ($s_t = 4$)}} &
    \!\!\!\!\includegraphics[width=0.9\linewidth]{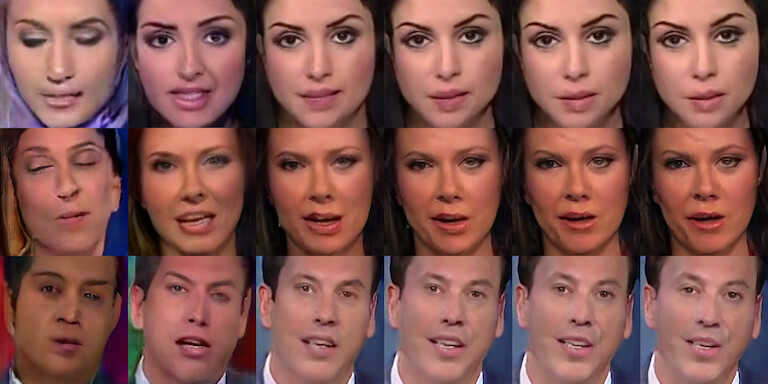} \\
    &\!\!\!\!\text{\small Frame 1 \hspace{50mm} Frame 6} \\
\end{tabular}
\caption{Comparison of instability of first frames trained with FaceForensics. Only the initial six frames are shown. }
\label{fig:face_forensics_diff}

\end{figure}

On the other hand, videos generated by the model of $s_t=4$ tended to be inconsistent in the first few frames
(note that we confirmed that both $s_t=2$ and $s_t=4$ generate stable videos for the remaining frames).
This example is shown in \Fig{face_forensics_diff}.
We considered it is because that the increase in the number of subsampled frames
leads to instability in the first few frames.
We also confirmed that this behavior depends on the dataset.
Specifically, the instability of $s_t=4$ has been mitigated when trained with UCF101 dataset, and we observed that in the quantitative evaluation, the model with $s_t=4$ outperformed the model with $s_t=2$.
We describe the detail in Sections \ref{sec:qualitative_ucf101} and \ref{sec:frame_subsampling}.

\subsubsection{UCF101}
\label{sec:qualitative_ucf101}

Next, we performed a similar experiment using UCF101,
which is more challenging to generate videos that look more natural than FaceForensics.
Unlike FaceForensics with $256 \times 256$ pixels,
the resolution of the UCF101 used in experiments is $192 \times 192$ pixels;
this is because the resolution of the original video is $320 \times 240$ pixels.
Four GPUs were used to train models.
Similar to the experiment in the FaceForensics, we trained two models of $s_t=2$ and $s_t=4$ with the UCF101 dataset.
The total batch size was set to 32, which is identical to the experiment in the FaceForensics.
To confirm the differences in qualitative results when using a conditional model,
we also trained conditional models with discrete labels in UCF101.
These hyperparameters are identical to those without conditions.

Although the metrics such as Inception Score and Fr\'echet Inception Distance used in
this quantitative experiment have been applied in the field of video generation
\cite{Saito2017,Tulyakov2018,Unterthiner2018},
it is difficult to know how much improvement in these values leads to qualitative improvement.
To confirm this difference, we compared qualitatively with videos generated by five models.
One is a TGAN \cite{Saito2017} trained with UCF101.
The second and third are our proposed models with $s_t=2$ and $s_t=4$.
The rest are the two baseline models described in \Sec{baselines}.
For the MoCoGAN \cite{Tulyakov2018},
we could not use it for the qualitative experiment
because the pre-trained model was not available and
there is no figure representing the generated result in the paper.
However, we considered that TGAN can also be used to compare the qualitative quality of
the current state-of-the-art methods because the TGAN achieves similar inception score to MoCoGAN.

\begin{figure*}
\begin{tabular}{ccc}
    \!\!\!\!\includegraphics[width=0.33\linewidth]{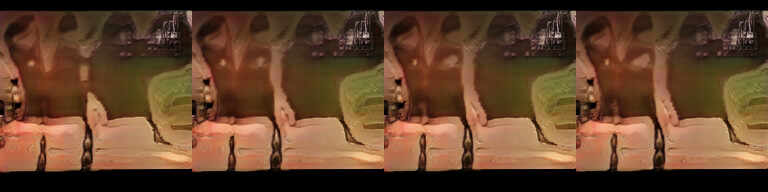} &
    \!\!\!\!\includegraphics[width=0.33\linewidth]{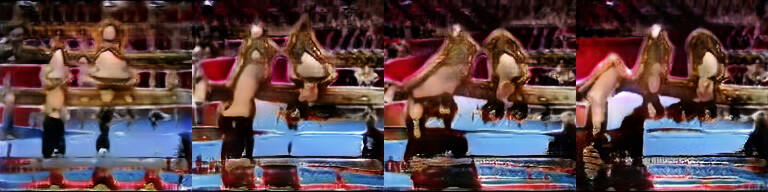} &
    \!\!\!\!\includegraphics[width=0.33\linewidth]{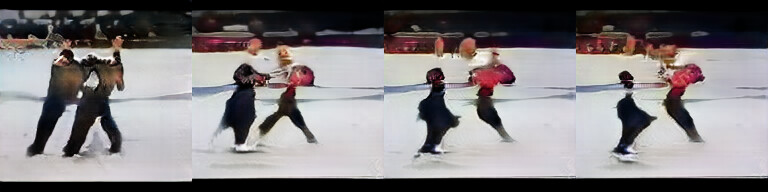} \\
    \!\!\!\!\includegraphics[width=0.33\linewidth]{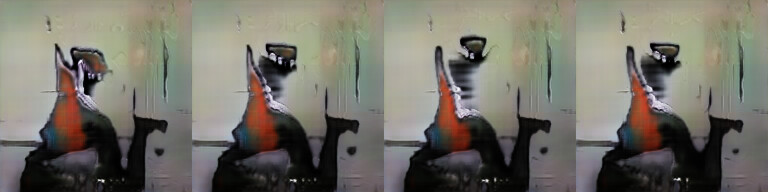} &
    \!\!\!\!\includegraphics[width=0.33\linewidth]{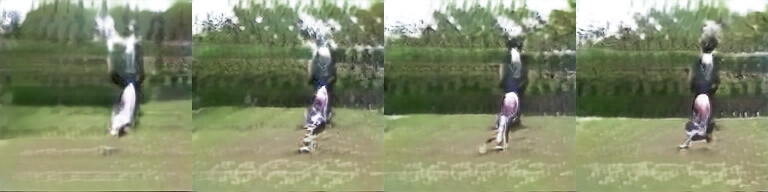} &
    \!\!\!\!\includegraphics[width=0.33\linewidth]{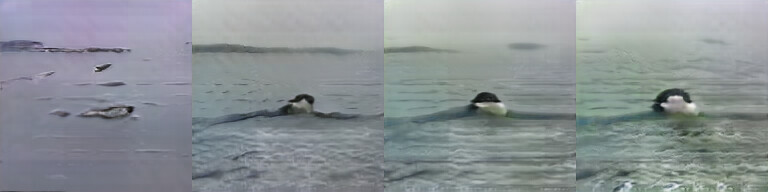} \\
    \!\!\!\!\includegraphics[width=0.33\linewidth]{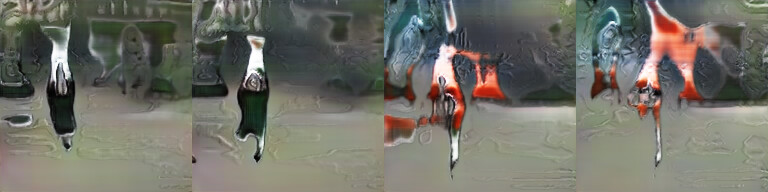} &
    \!\!\!\!\includegraphics[width=0.33\linewidth]{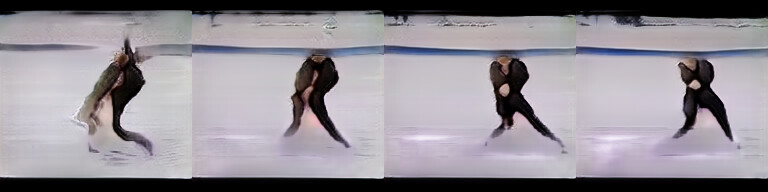} &
    \!\!\!\!\includegraphics[width=0.33\linewidth]{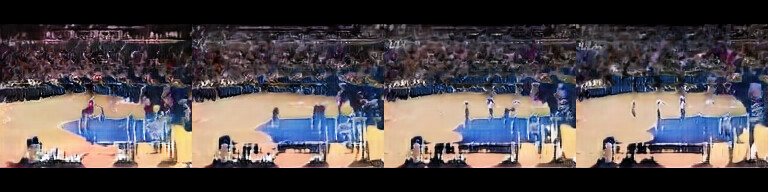} \\
    \!\!\!\!\includegraphics[width=0.33\linewidth]{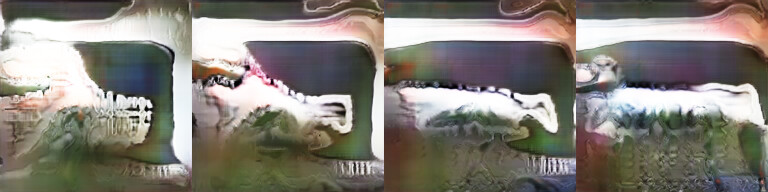} &
    \!\!\!\!\includegraphics[width=0.33\linewidth]{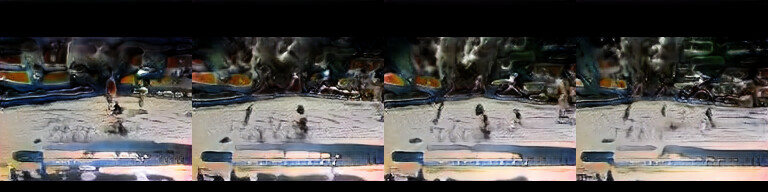} &
    \!\!\!\!\includegraphics[width=0.33\linewidth]{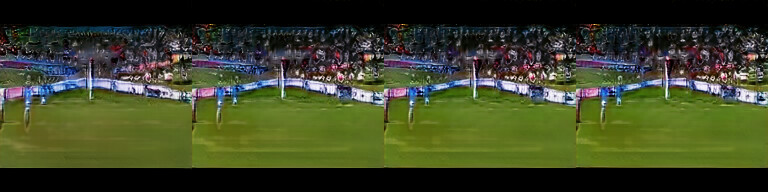} \\
    \!\!\!\!\text{\small Frame 1 \hspace{30mm} Frame 16} &
    \!\!\!\!\text{\small Frame 1 \hspace{30mm} Frame 16} &
    \!\!\!\!\text{\small Frame 1 \hspace{30mm} Frame 16} \\
    \!\!\!\!\text{\small (i) 3D + 2D discriminators} &
    \!\!\!\!\text{\small (ii) Our model ($s_t = 2$, uncond.)} &
    \!\!\!\!\text{\small (iii) Our model ($s_t = 4$, uncond.)} \\
\end{tabular}
\caption{
Example of videos by the three models (our two unconditional models ($s_t=2, 4$) and baseline (``3D + 2D discriminators'')) trained with UCF101.
Images excluding the intermediate three frames are shown to make it easy to identify the motion.}
\label{fig:ucf101}

\end{figure*}

\begin{figure*}
\begin{tabular}{cc}
    \!\!\!\!\includegraphics[width=0.5\linewidth]{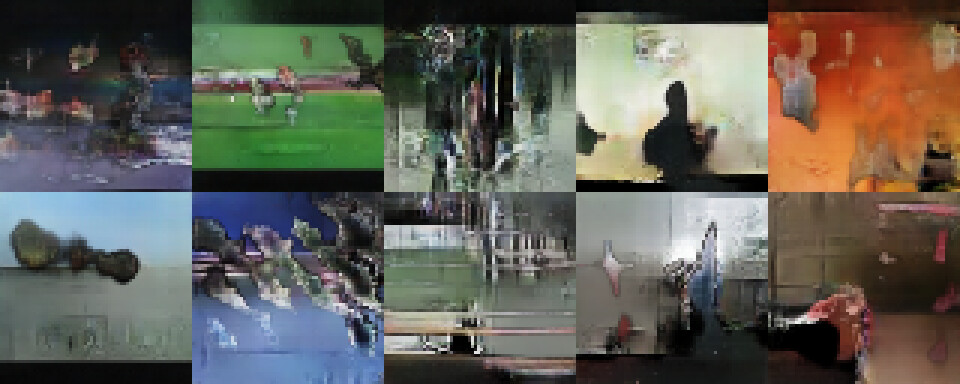} &
    \!\!\!\!\includegraphics[width=0.5\linewidth]{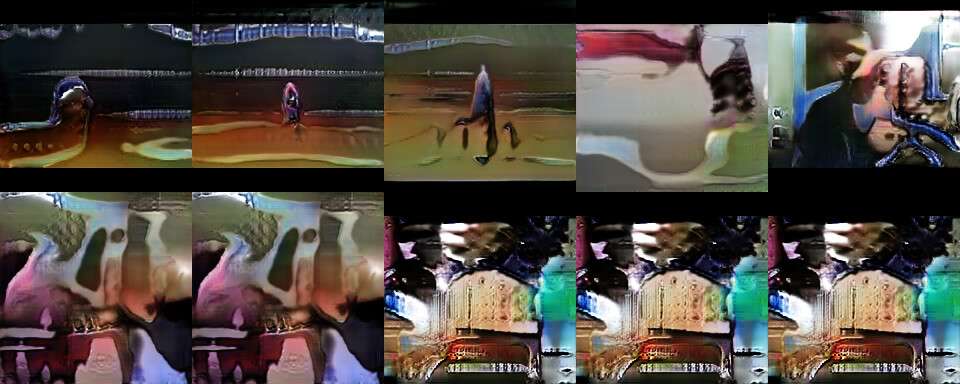} \\
    \!\!\!\!\text{\small TGAN \cite{Saito2017}} &
    \!\!\!\!\text{\small Single 3D discriminator} \\
    \!\!\!\!\includegraphics[width=0.5\linewidth]{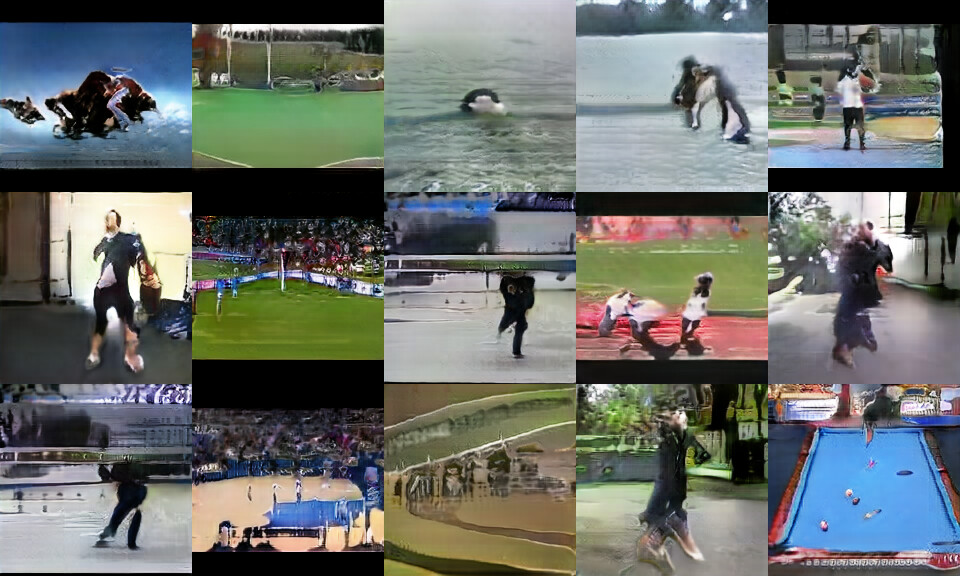} &
    \!\!\!\!\includegraphics[width=0.5\linewidth]{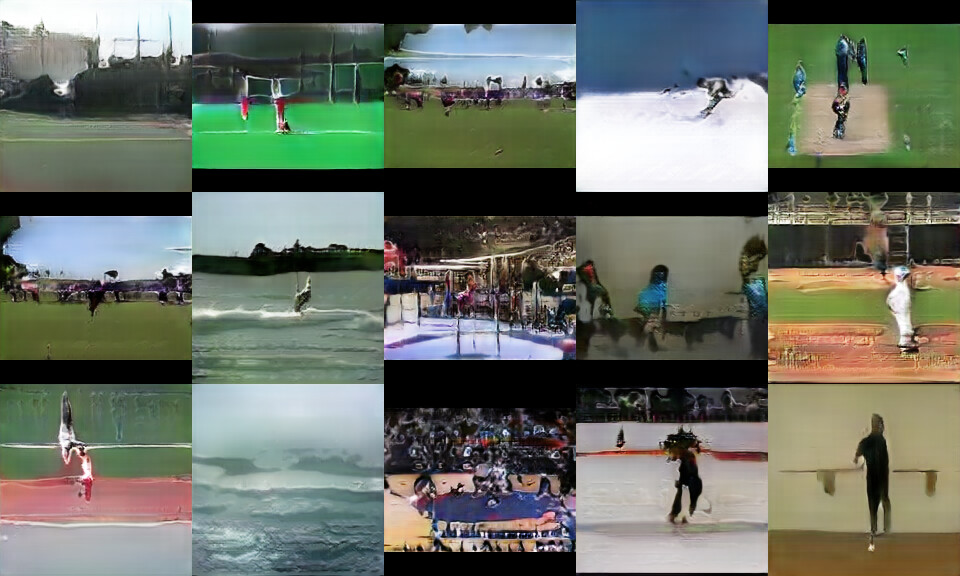} \\
    \!\!\!\!\text{\small Our model ($s_t = 4$, unconditional)} &
    \!\!\!\!\text{\small Our model ($s_t = 4$, conditional)} \\
\end{tabular}
\caption{
A list of still images extracted from videos
generated by our models ($s_t=4$), TGAN, and baseline model.
The UCF101 dataset was used for the training.
``unconditional`` means a generator trained with videos only,
whereas ``conditional'' is a conditional one trained with videos and corresponding labels.
}
\label{fig:ucf101_list}
\end{figure*}

\Fig{ucf101} shows the videos generated by the baseline and our two models.
We confirmed that our proposed two models generated videos that are easier to interpret the content than the two baseline models.
Specifically, the videos generated by our models tend to have some abstract motions such as moving shadows that look like a human, but its outline is relatively clear, and the background is more distinguishable (e.g., indoor, sea, grassland, and stadium) than baseline models.
In particular, when the background of the video is static such as soccer or basketball in a distant view, our model generated more detailed videos than other conventional models.

The baseline models trained with UCF101 tended to
produce the same meaningless videos since they often caused mode collapse.
On the other hand, we did not find such a tendency in the proposed models.
In order to show this behavior, we list the still images
extracted from the generated videos in \Fig{ucf101_list}.
The inception scores and Fr\'echet inception distance measured by these models
are much better than those observed by the baseline models.
It suggests that both metrics have a certain degree of correlation with the quality of videos perceived by humans even in UCF101.
For reference, we also put a list of frames extracted from the videos generated by TGAN \cite{Saito2017}.
We also confirmed that the TGAN tends to generate noisy videos and blurry backgrounds,
whereas our model tends to produce a video with a relatively clear background and sharp shadows.

As for videos generated by the condtional model, we observed that it tends to yield a relatively stable video than the unconditional one.
This was especially noticeable when the content of the video was a distant view.
We have also confirmed through quantitative experiments described later that the quality of the videos by the conditional model
tends to exceed that of the unconditional one.
Because the quality of the video in which a person moves well is still comparable to the unconditional model,
we inferred that such quantitative improvement of the quality was mainly because of the improvement of distant view videos.

\begin{figure}
\begin{tabular}{cc}
    \!\!\!\!\rotatebox{90}{\parbox{3.0cm}{Our model ($s_t = 2$)}} &
    \!\!\!\!\includegraphics[width=0.9\linewidth]{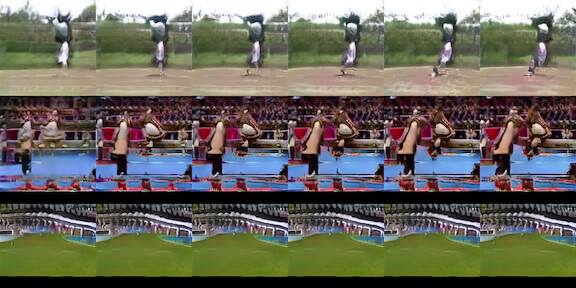}\\
    \!\!\!\!\rotatebox{90}{\parbox{3.0cm}{Our model ($s_t = 4$)}} &
    \!\!\!\!\includegraphics[width=0.9\linewidth]{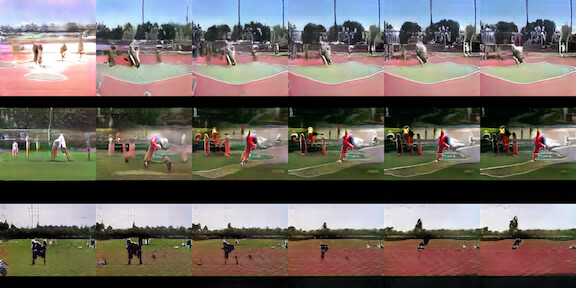} \\
    &\!\!\!\!\text{\small Frame 1 \hspace{50mm} Frame 6} \\
\end{tabular}
\caption{Comparison of instability of first frames trained with UCF101. Only the initial six frames are shown.}
\label{fig:ucf101_diff}
\end{figure}

Unlike the experiment of FaceForensics with $s_t=4$,
in the experiment of UCF101 with $s_t=4$
we did not see such steep instability but observed gradual changes in the video,
i.e., the category of the video does not change but topography changes gradually.
The example is shown in \Fig{ucf101_diff}.
We considered that it was due to differences in the domain of the dataset.
If videos in the dataset are not very diverse as in FaceForensics,
the transition of the domain of the first frames caused by the subsampling layer tends to occur.
Contrarily, if the dataset is diverse as in UCF101,
we considered that such steep transition would be less likely to occur since such transitions are easily identified by the discriminator.

\subsection{Linear interpolation of the noise vector}
\begin{figure}[t]
\begin{tabular}{cccccc}
    \!\!\!\!\includegraphics[width=0.14\linewidth]{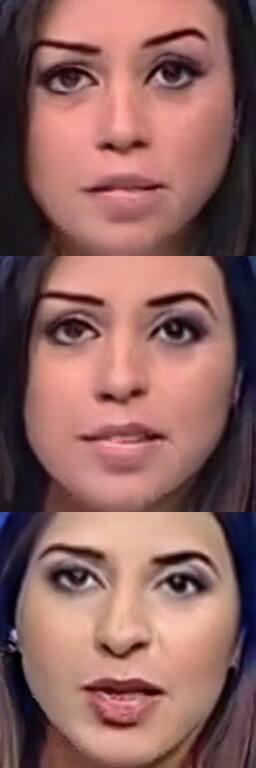} &
    \!\!\!\!\includegraphics[width=0.14\linewidth]{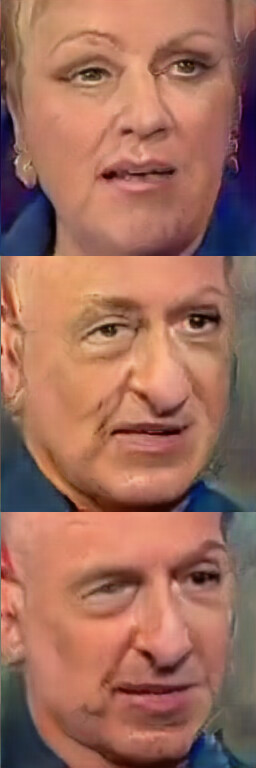} &
    \!\!\!\!\includegraphics[width=0.14\linewidth]{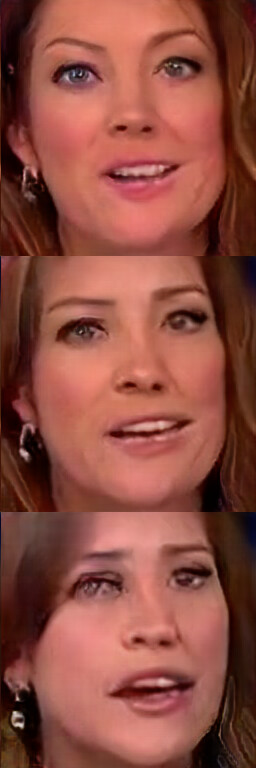} &
    \!\!\!\!\includegraphics[width=0.14\linewidth]{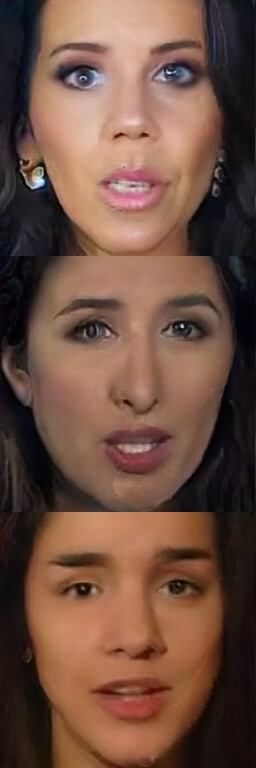} &
    \!\!\!\!\includegraphics[width=0.14\linewidth]{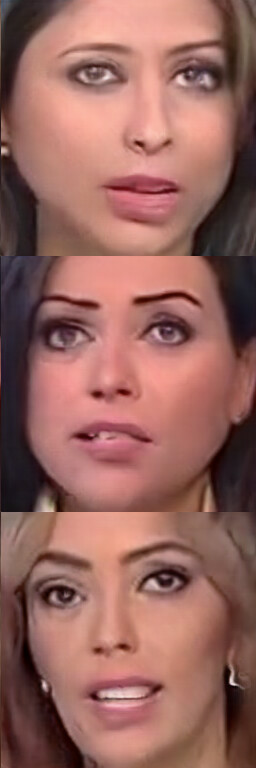} &
    \!\!\!\!\includegraphics[width=0.14\linewidth]{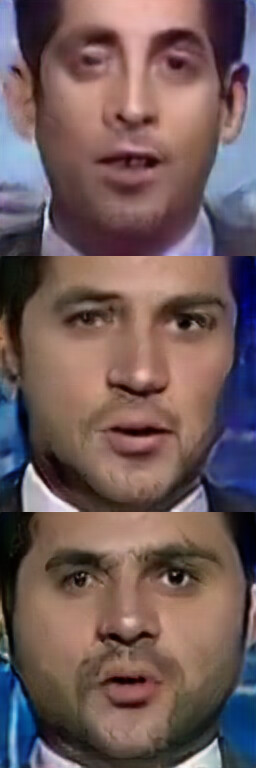} \\
\end{tabular}
\caption{Example of linear interpolation of noise vectors. The first and last rows represent images sampled from the two noise vectors, and the middle row means an image generated from an intermediate vector between them. Only the result of the eight frame is shown for simplicity. The model trained with $s_t = 2$ was used.}
\label{fig:interpolation}
\end{figure}

We also generated a new sample with a intermediate vector between the two noise vectors
to ensure that the trained network was not a model that memorized the dataset.
The model trained with $s_t = 2$ and the FaceForensics dataset was used for the experiment.

The result is shown in \Fig{interpolation}.
The content of the video changed smoothly by moving $\vec{z}$ linearly.
It means that our model is not just a ``memorized'' model of the original dataset.
Changing $\vec{z}$ often tended to change the speaker itself, but on the other hand,
there were several videos where the speaker did not change,
but other attributes such as the orientation of the speaker's face changed smoothly.
We inferred that it was mainly because that the input of the discriminator
was a sequence clipped at 16 frames randomly from a video in the dataset.

\subsection{Consistency of generated samples}
\begin{figure}[t]
\begin{tabular}{cc}
    \!\!\!\!\includegraphics[width=0.47\linewidth]{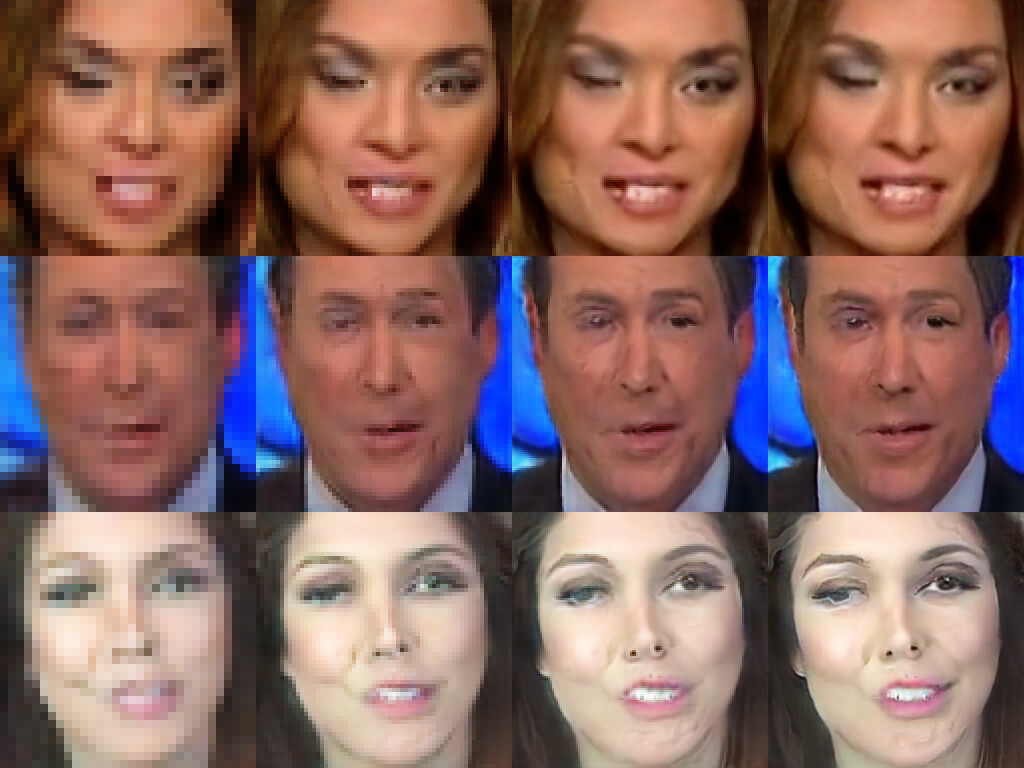} &
    \!\!\!\!\includegraphics[width=0.47\linewidth]{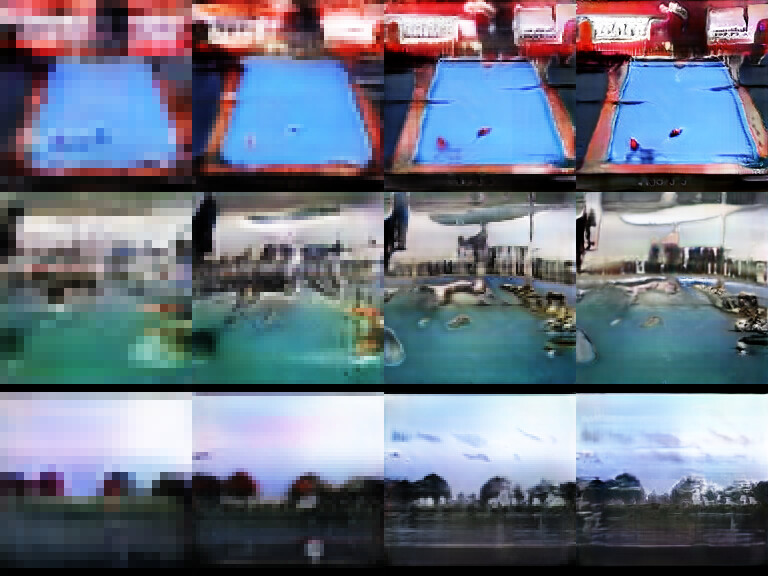} \\
    \!\!\!\!\text{\small Level 1 \hspace{18mm} Level 4} &
    \!\!\!\!\text{\small Level 1 \hspace{18mm} Level 4} \\
    \!\!\!\!\text{\small (i) FaceForensics ($s_t=2$)} &
    \!\!\!\!\text{\small (ii) UCF101 ($s_t=4$)} \\
\end{tabular}
\caption{Example of videos generated at different levels.
The leftmost image represents the frame of the video
generated by the initial rendering block,
and the rightmost one denotes the final generated image.}
\label{fig:transition}
\end{figure}
To check whether each rendering block generates the same content under the same $\vec{z}$, we visualized the result of each rendering block. The example is shown in \Fig{transition}.
We observed that every rendering block in our method tends to generate the same content without a color-consistency regularization introduced in StackGAN++ \cite{Zhang2017a}.

\begin{figure}[t]
\begin{tabular}{ccc}
    \!\!\!\!\rotatebox{90}{\parbox{3.8cm}{\small Level 1 \hspace{15mm} Level 4}} &
    \!\!\!\!\includegraphics[width=0.45\linewidth]{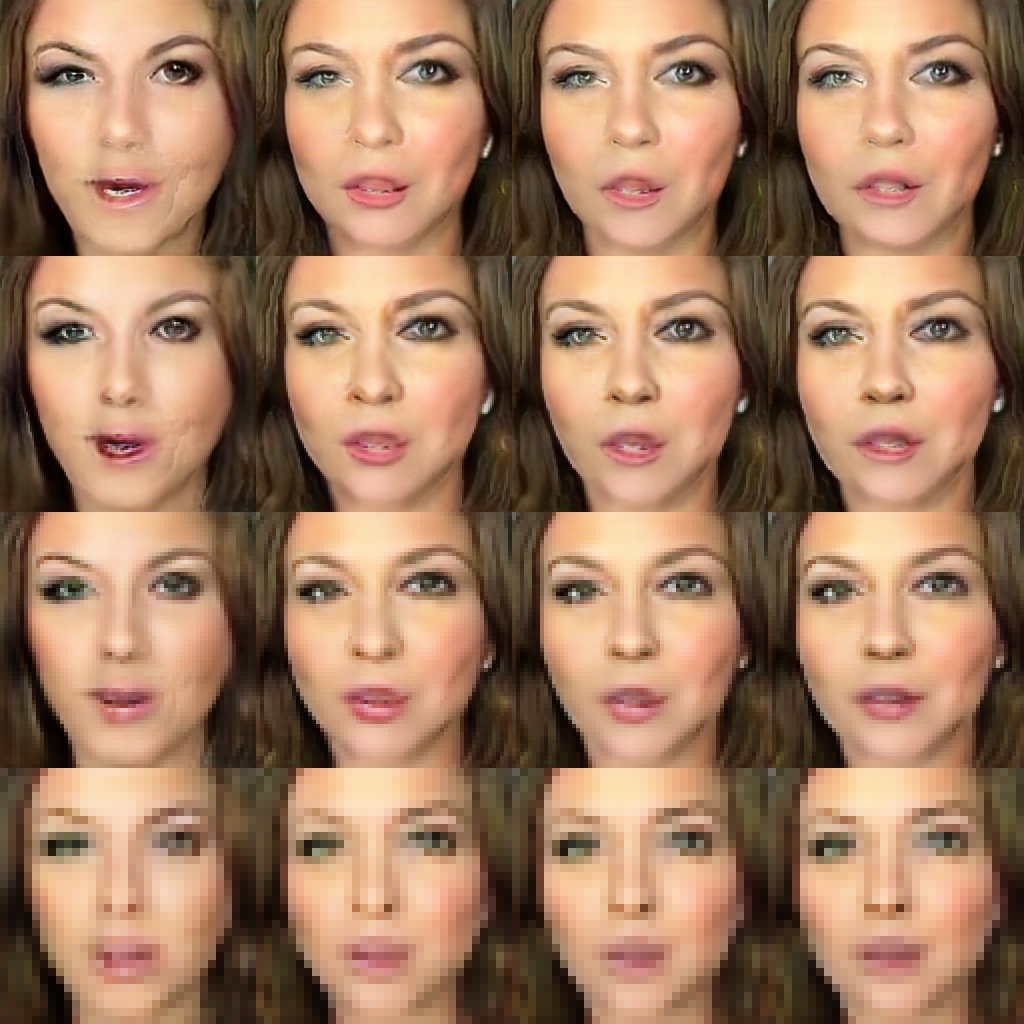} &
    \!\!\!\!\includegraphics[width=0.45\linewidth]{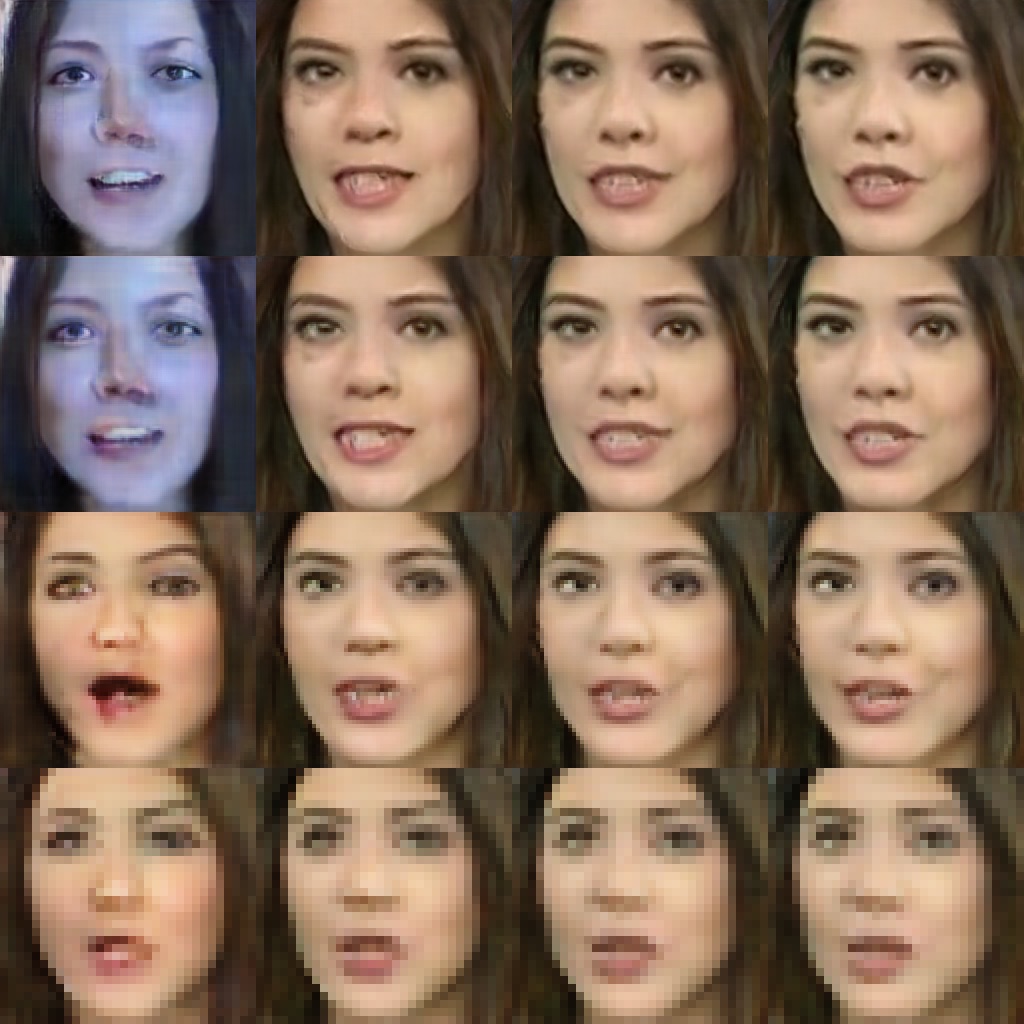} \\
    \!\!\!\!&
    \!\!\!\!\text{\small Frame 1 \hspace{12mm} Frame 4} &
    \!\!\!\!\text{\small Frame 1 \hspace{12mm} Frame 4} \\
    \!\!\!\!&
    \!\!\!\!\text{\small $s_t=2$} &
    \!\!\!\!\text{\small $s_t=4$} \\
\end{tabular}
\caption{The difference of the video which each rendering block outputs. The row represents levels of rendering blocks and time, and the column means time.}
\label{fig:multilevel_difference}
\end{figure}

Next, making use of the property that all blocks tend to output the same content,
we observed which level of blocks the instability of the first frames at $s_t=4$ resulted from.
To check the results of the videos generated by all rendering blocks,
we disabled every subsampling layer to make the rendering block of each level output a video with 16 frames.
The results for each level at $s_t=2, 4$ are shown in \Fig{multilevel_difference}.
In contrast to the result of $s_t=2$, where every rendering block generated a video with almost identical content, the first frames of the model of $s_t=4$ tended to become unstable as the level increased.
It indicates that the cause of instability is that the subsequent rendering blocks no longer output the same content.

\begin{figure*}[t]
\begin{tabular}{cc}
    \!\!\!\!\includegraphics[width=0.5\linewidth]{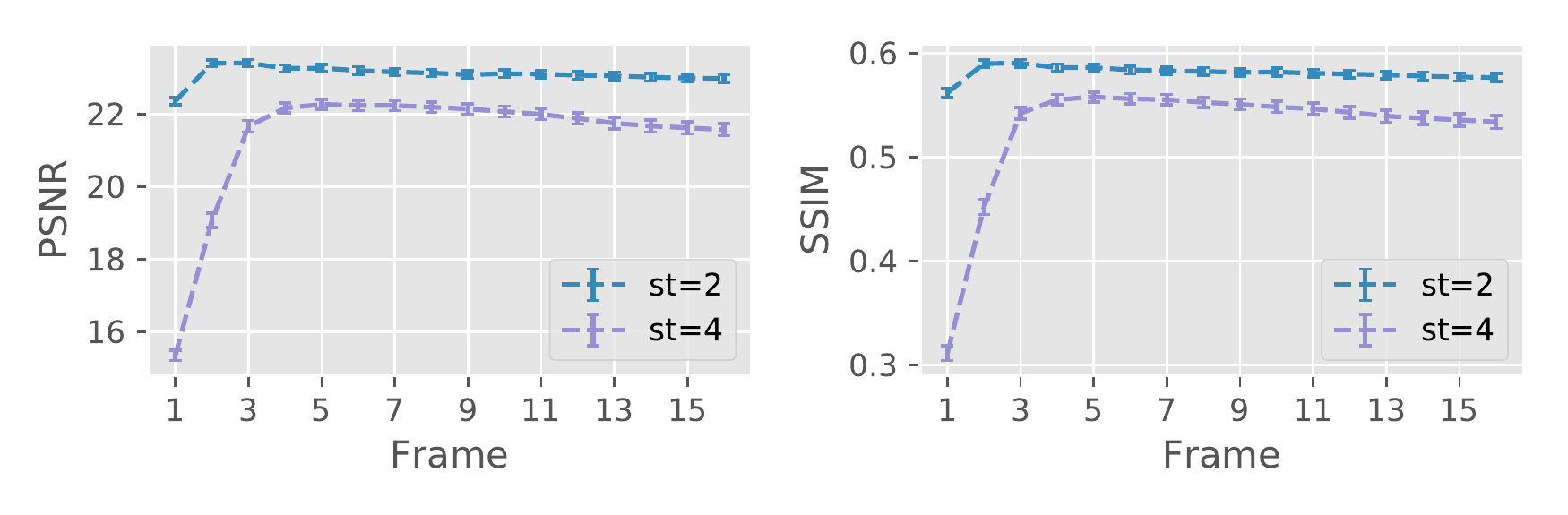} &
    \!\!\!\!\includegraphics[width=0.5\linewidth]{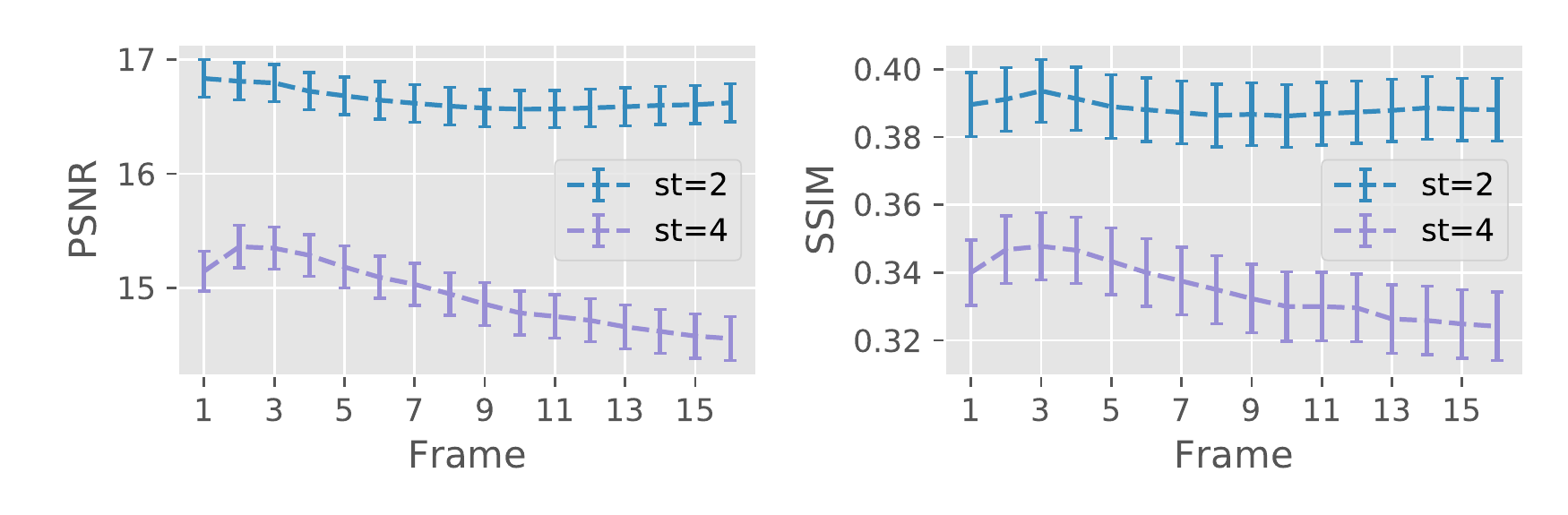} \\
    \!\!\!\!\text{\small (i) FaceForensics} &
    \!\!\!\!\text{\small (ii) UCF101} \\
\end{tabular}
\caption{
Quantitative differences of the videos generated by different rendering blocks.
The row indicates the difference between the two frames generated by rendering block at level 1 and 4.
The column means time.
Error bars at the 95\% confidence interval.}
\label{fig:multilevel_psnr_ssim}
\end{figure*}

To quantitatively confirm that this instability depends on the dataset,
we calculated PSNR (Peak Signal-to-Noise Ratio) and SSIM (structural similarity) between
the frame in a video generated by the rendering block of level 1 and the frame of level 4.
We resized the videos in level 1 so that the resolutions of the two videos match.
One thousand samples were used for evaluation.
These quantitative results are shown in \Fig{multilevel_psnr_ssim}.
It indicates that the first frames at level 4 become unstable when
trained with the FaceForensics dataset, while later it outputs almost identical content as level 1.
However, the identity of $s_t=4$ was worse than $s_t=2$ even if enough time had passed.
While such a rapid initial instability was not observed with UCF101,
its identity after a sufficient period of time was slightly worse for both SSIM and PSNR than for FaceForensics.
It means that the identity of the video generated by each block changes depending on the difficulty level of the dataset.

\subsection{Comparison with other existing methods}
\label{sec:exp_comparison}
We confirmed that the quality of samples generated by the proposed model is superior to other existing models with Inception Score (IS) \cite{Salimans2016} and Fr\'echet Inception Distance (FID) \cite{Heusel2017}, which measure the quality of generated samples. We describe the details of the experimental environment used for measuring IS and FID in the following. See \cite{Borji2018} for details of IS and FID.

As with other comparable methods \cite{Saito2017,Tulyakov2018,Acharya2018},
we employed the UCF101 dataset \cite{Soomro2012} for the quantitative experiments.
Regarding the classifier used for computing IS and FID,
we used a pre-trained model of Convolutional 3D (C3D) network \cite{Tran2015},
which was first trained with Sports-1M dataset \cite{Karpathy2014}
and then fine-tuned on the UCF101 dataset\footnote{
The pre-trained model itself can be downloaded from \url{http://vlg.cs.dartmouth.edu/c3d/}}.
Note that we used this pre-trained model as is,
and did not update any parameters in the classifier.
As the resolution and the number of frames in a video used in the classifier are
$128 \times 128$ pixels and 16 frames,
we resized the generated video with $192 \times 192$ pixels to $128 \times 128$ pixels
and regarded it as the input of the classifier. 

To calculate the IS and FID with the C3D model, we need a fourth-order tensor ($C \times T \times H \times W$) that represents the average for all the clips in the UCF101.
We used the same average tensor used in TGAN.

In all experiments, we computed FID and IS according to the following procedure.
First, we set the maximum number of iterations to 100,000 and took snapshots every 2,000 iterations. For each snapshot, we computed the IS and FID with 2,048 samples.
The best snapshot with the largest IS was used for quantitative evaluation.
We calculated the mean and standard deviation by performing the same procedure ten times for this snapshot.
It took about four hours to compute IS and FID of one snapshot.

The resolutions of generated videos by all the existing methods are $64 \times 64$ pixels
except for ProgressiveGAN ($128 \times 128$ pixels).
These scores shown in the quantitative experiments in these models can be compared with
the IS calculated by our method because their models also compute the IS
by resizing the generated videos to $128 \times 128$ pixels.
However, even if the IS of our method exceeds the existing method,
this increase may be due to simply increasing the resolution
(in other words, the IS may be an indicator with a vulnerability that can be easily increased by increasing resolution).
To confirm this, we measured IS and FID of two baseline models
with a similar network structure as the existing method (c.f., \Sec{baselines}),
and observed the change of both scores in the case of high resolution.
The difference between our model and the existing ones is roughly divided into the two:
multi-scale model and subsampling layer.
To clarify the contribution of each factor to the IS and FID,
we also defined a model without all subsampling layers as
``naive implementation'', and measured its IS and FID.

Four GPUs were used in all the experiments of this subsection.
We set the maximum batch size within the memory of the GPU for each model.
This means that the batch size of the two proposed methods was 32,
whereas that of two baseline models was 8.
The model without subsampling layers consumes much memory due to the resolution;
therefore, we had to set the entire batch size to 4 even if we can use four GPUs.

\begin{table}[t]
\centering
{\renewcommand{\arraystretch}{1.2}
\begin{tabular}{lll}
Method & IS & FID \\ \hline \hline
VGAN \cite{Vondrick2016} & $8.31 \pm .09$ \\
TGAN \cite{Saito2017} & $11.85 \pm .07$ \\
MoCoGAN \cite{Tulyakov2018} & $12.42 \pm .03$ \\
ProgressiveVGAN \cite{Acharya2018} & $13.59 \pm .07$ \\
ProgressiveVGAN w/ SWL \cite{Acharya2018} & $14.56 \pm .05$ \\ \hline
Single 3D discriminator only & $11.10 \pm .16$ & $8358 \pm 81$ \\
3D + 2D discriminators & $10.47 \pm .12$ & $8304 \pm 70$ \\
Naive implementation & $13.29 \pm .15$ & $5401 \pm 31$ \\ \hline

Our model ($s_t = 2$) & $23.87 \pm .28$ & $3797 \pm 20$ \\
Our model ($s_t = 4$) & $26.60 \pm .47$ & $3431 \pm 19$ \\ \hline
\end{tabular}
}
\caption{Inception Score and Fr\'echet Inception Distance on UCF101 dataset.}
\label{table:inception_score}
\vspace{-10pt}
\end{table}

The quantitative results are shown in \Table{inception_score}.
The inception scores of two baseline models are slightly lower than those of other existing methods.
It indicates that it is difficult to improve scores
by simply increasing the resolution of the samples from the generator.
Although it may be possible that the baseline models could not allocate a sufficient mini-batch
due to the high resolution, we will show in \Sec{exp_comparison_same_bs} that
these scores do not increase dramatically even if the batch size is sufficiently large.
The IS and FID of the baseline model with two sub-discriminators are almost the same
as those of the baseline with a single 3D discriminator.
It implies that it is difficult to improve the quality of the generated videos dramatically
with the conventional approach of simply introducing multiple discriminators into the input layer.
The IS and FID of the model without subsampling layers are much lower than those of our proposed model.
It indicates that the subsampling layer contributes to the improvement of the quality
in an environment of four GPUs.

The inception scores of our two models ($s_t=2, 4$) are
significantly higher than those of the other existing models.
In other words, in the environment of four GPUs,
it means that the quality of our proposed models exceeds the existing method.
It is interesting to see that IS and FID computed by the model of $s_t=4$
are better than those of $s_t=2$.
Thus, by increasing the number of frames to be reduced,
our method can not only save the computational cost and the GPU memory more efficiency
but also slightly improve the quality of the generated videos.
As discussed in Sections \ref{sec:face_forensics} and \ref{sec:qualitative_ucf101},
increasing $s_t$ may contribute to instability of first few frames in some datasets;
however, the increase of IS by setting $s_t=2$ instead of not using subsampling layers
is much higher than the increase of IS by setting $s_t=4$ instead of $s_t=2$.
Based on these quantitative results, we concluded that an approach of setting
$s_t=2$ at the beginning and then gradually increasing would be appropriate
when training with an unknown dataset.

\subsection{Quantitative results under a single GPU}
\label{sec:single_gpu_exp}
\begin{table}[t]
\centering
{\renewcommand{\arraystretch}{1.2}
\begin{tabular}{lll}
Method & IS & FID \\ \hline \hline
Single 3D discriminator only & $3.41 \pm .03$ & $13252 \pm 50$ \\
3D + 2D discriminators & $1.01 \pm .01$ & $15380 \pm 30$ \\
Naive implementation & $1.43 \pm .01$ & $13698 \pm 12$ \\ \hline
Our model ($s_t = 2$) & $20.61 \pm .28$ & $3930 \pm 25$ \\
Our model ($s_t = 4$) & $21.45 \pm .29$ & $3877 \pm 15$ \\ \hline
\end{tabular}
}
\caption{Inception Score and Fr\'echet Inception Distance when using a single GPU.}
\label{table:is_single_gpu}
\vspace{-10pt}
\end{table}

Although our proposed model outperforms the other existing methods
in the environment of 4 GPUs, this improvement may be due to
the advantageous setting where we used multiple GPUs.
To confirm whether the IS and FID of our method exceed
those of the existing method under the almost same environment,
we measured IS and FID of the two baselines and the proposed models
when training with a single GPU.

The results are shown in \Table{is_single_gpu}.
IS and FID of the model without subsampling layers are much worse than
those of the existing models.
It is because that the video is a high resolution and the batch size is only one.
It also illustrates the difficulty of training high resolution models
in a limited resource.
Even if we can use only one GPU, the IS of our model is
significantly higher than those of the other existing models.
In particular, the fact that inception scores of our models
are higher than that of ProgressiveGAN,
which also outputs high-resolution videos,
demonstrates the effectiveness of our model under constrained computational resources.

\subsection{Quantitative results under the same batch size}
\label{sec:exp_comparison_same_bs}
\begin{table}[t]
\centering
{\renewcommand{\arraystretch}{1.2}
\begin{tabular}{lll}
Method & IS & FID \\ \hline \hline
Single 3D discriminator only & $11.72 \pm .20$ & $7754 \pm 47$ \\
3D + 2D discriminators & $13.38 \pm .22$ & $6993 \pm 25$ \\
Naive implementation & $14.44 \pm .20$ & $7613 \pm 29$ \\ \hline
Our model ($s_t = 2$) & $23.87 \pm .28$ & $3797 \pm 20$ \\
Our model ($s_t = 4$) & $26.60 \pm .47$ & $3431 \pm 19$ \\ \hline
\end{tabular}
}
\caption{Inception Score and Fr\'echet Inception Distance when setting 32 for batch size.}
\label{table:is_bs_32}
\vspace{-10pt}
\end{table}

It can be seen that our models outperformed the other existing and baseline models
under the environment of one or four GPUs.
However, if we can increase the batch size without such limitations,
i.e., if we can set the same batch size as our model by simply increasing the number of GPUs,
baseline models may outperform the proposed method.
Therefore, we measured IS and FID of the three baseline models under the same batch size.
The batch size was set to 32. It means that the model without the subsampling layers consumed 32 GPUs, whereas the two baseline models used 16 GPUs and our method used only four GPUs.

\Table{is_bs_32} shows the results. Even if there is no limit on the number of GPUs,
our proposed model outperformed all the baseline models.
We can claim the following two from these results.
First, the low IS and the high FID in the baseline models shown in the previous experiments
are not mainly due to the small batch size.
That is, it is hard to improve both scores dramatically
with a simple model that introduces a single discriminator only to the input layer.
Second, the reason why both scores of our models are much better than those of the existing methods
is not a simple reason like the number of parameters is larger than others.
Even in an environment where a sufficient number of GPUs can be used,
the IS of the model without introducing the subsampling layer was similar to the other existing methods.
It implies that such high scores are because of the introduction of multiple subsampling layers.

\subsection{Effectiveness of frame sub-sampling layers}
\label{sec:frame_subsampling}

\begin{table}
\centering
{\renewcommand{\arraystretch}{1.2}
\begin{tabular}{c|ccc}
Batchsize & Naive impl. & $s_t = 2$ & $s_t = 4$ \\ \hline \hline
4 & {$13.29 \pm .15$} & {$14.64 \pm .25$} & ${17.12 \pm .26}$ \\
8 & {$11.26 \pm .19$} & {$20.61 \pm .28$} & ${21.45 \pm .29}$ \\
16 & {$12.79 \pm .10$} & {$22.81 \pm .52$} & ${22.69 \pm .47}$ \\
32 & {$14.44 \pm .20$} & {$23.87 \pm .28$} & ${26.60 \pm .47}$ \\
64 & N/A & {$24.39 \pm .50$} & ${26.89 \pm .40}$ \\
128 & N/A & {$25.65 \pm .34$} & {$28.87 \pm .67$} \\ \hline
\end{tabular}
}
\caption{Changes in the inception score when sub-sampling layers are enabled.}
\label{table:fr_is}
\end{table}

\begin{table}
\centering
{\renewcommand{\arraystretch}{1.2}
\begin{tabular}{c|ccc}
Batchsize & Naive impl. & $s_t = 2$ & $s_t = 4$ \\ \hline \hline
4 & {$5401 \pm 31$} & ${5148 \pm 33}$ & ${4449 \pm 14}$ \\
8 & {$7364 \pm 51$} & {$3930 \pm 25$} & ${3877 \pm 15}$ \\
16 & {$7822 \pm 27$} & {$3639 \pm 15$} & ${3715 \pm 19}$ \\
32 & {$7613 \pm 29$} & {$3797 \pm 20$} & ${3431 \pm 19}$ \\
64 & N/A & ${4242 \pm 32}$ & ${3334 \pm 26}$ \\
128 & N/A & {$3573 \pm 28$} & {$3497 \pm 26$} \\ \hline
\end{tabular}
}
\caption{Changes in the Fr\'echet Inception Distance when sub-sampling layers are enabled.}
\label{table:fr_fid}
\end{table}

IS and FID highly depend on batch size.
We confirmed that activating the subsampling layer improved IS and FID of our model over the other existing methods,
and the scores of the model of $s_t=4$ are slightly better than those of $s_t=2$.
However, it is not clear whether these trends are the same even if the batch size is increased
(for example, the score of the naive model may catch up with that of the model with subsampling layers).
To see the effectiveness of the subsampling layer for batch size,
we measured the change of the inception score when the subsampling layers were activated
under the condition of the same batch size.
In this subsection we performed three types of experiments;
one is an experiment when all subsampling layers are disabled,
and the remaining two are experiments when the subsampling layers are enabled with $s_t=2$ and 4.
Training these models with different batch sizes,
we measured the change of IS and FID.
For economic reasons, we did not measure scores of batch size 64 and 128
when disabling the subsampling layer.

The results are shown in Tables \ref{table:fr_is} and \ref{table:fr_fid}.
In any batch size, IS and FID of the model with subsampling layers
are significantly better than the model without them.
It indicates that the subsampling layers are effective for improving the quality of the generated videos
regardless of the batch size.
As the discriminator at the later stage has to determine the authenticity of the video with a lower frame rate,
it needs to judge the authenticity from the global motion and
the quality of a still image instead of the difference of the fine movement.
We considered that the subsampling layer played a role like {\em regularization},
resulting in improvement of IS and FID.

We also confirmed that IS and FID at $s_t=4$ were significantly
higher than those at $s_t=2$ in almost all batch sizes.
In particular, when the batch size is 128 and $s_t=4$,
the inception score of our model (28.87) dramatically exceeds
the existing state-of-the-art method (14.56).
It shows the superiority of our model over other existing methods.
Note that our IS also exceeds all the scores of the existing methods
when using one GPU only instead of
such relatively large computational resources
(c.f., \Sec{single_gpu_exp}).

\subsection{Quantitative results of the conditional model}
\begin{table}
\centering
{\renewcommand{\arraystretch}{1.2}
\begin{tabular}{c|cc|cc}
& \multicolumn{2}{|c|}{$s_t = 2$} & \multicolumn{2}{|c}{$s_t = 4$} \\ \cline{2-5}
BS & pure & conditional & pure & conditional \\ \hline \hline
16 & {$22.81 \pm .52$} & {$40.59 \pm .94$} & ${22.69 \pm .47}$ & {$39.61 \pm .94$} \\
32 & {$23.87 \pm .28$} & {$48.00 \pm 1.0$} & ${26.60 \pm .47}$ & {$49.30 \pm .56$} \\
64 & {$24.39 \pm .50$} & {$54.93 \pm .68$} & ${26.89 \pm .40}$ & {$51.21 \pm .49$} \\ \hline
\end{tabular}
}
\caption{Changes in the inception score when updating to a conditional model. ``pure`` denotes an unconditional model, whereas ``conditional`` is a conditional one described in \Sec{cgan}. ``BS'' means a batch size.}
\label{table:cond_is}
\end{table}

\begin{table}
\centering
{\renewcommand{\arraystretch}{1.2}
\begin{tabular}{c|cc|cc}
& \multicolumn{2}{|c|}{$s_t = 2$} & \multicolumn{2}{|c}{$s_t = 4$} \\ \cline{2-5}
BS & pure & conditional & pure & conditional \\ \hline \hline
16 & {$3639 \pm 15$} & {$3390 \pm 20$} & ${3715 \pm 19}$ & {$3487 \pm 22$} \\
32 & {$3797 \pm 20$} & {$3253 \pm 22$} & ${3431 \pm 19}$ & {$3186 \pm 16$} \\
64 & ${4242 \pm 32}$ & {$3934 \pm 24$} & ${3334 \pm 26}$ & {$3352 \pm 26$} \\ \hline
\end{tabular}
}
\caption{Changes in the Fr\'echet Inception Distance when updating to a conditional model.}
\label{table:cond_fid}
\end{table}

As we described in \Sec{cgan}, the quality of our generated video
can be improved with a conditional model.
To see the changes of quantitative scores when using the conditional model,
we measured the changes in IS and FID for each batch size.
The experiment was performed under two conditions: $s_t = 2$ and $s_t = 4$.
Three batch sizes of 16, 32, and 64 were used in the experiment.

The results are shown in Tables \ref{table:cond_is} and \ref{table:cond_fid}.
They show that the scores improved significantly by extending to the conditional model
regardless of the batch size and hyperparameters.
In particular, the inception score (54.93) of the conditional model for $s_t = 2$
and the batch size of 64 is significantly higher than any previously reported scores.
However, an increase in batch size did not lead to a significant increase in FID,
but was sometimes worsened.
When the batch size was 32, the FID of the conditional model was improved
over all scores in \Table{fr_fid},
but its improvement is not as dramatic as that of IS.
It is more intuitive when looking at the actual generated videos (see \Fig{ucf101_list}).

\subsection{Effectiveness of each subsampling layer}

In previous experiments, we enabled multiple subsampling layers simultaneously.
To confirm whether all of the subsampling layers introduced by our method
actually contribute to the improvement of the quality,
we measured the change of IS and FID
when multiple subsampling layers were gradually activated.

The concrete procedure is as follows.
First, according to the notation in \Fig{multigen}
we defined the three subsampling layers introduced in our method as
${\cal S}_1$, ${\cal S}_2$, and ${\cal S}_3$.
Next, we observed the change of IS and FID
when these layers were activated sequentially from ${\cal S}_1$.
Enabling ${\cal S}_i$ affects the number of video frames
received by all subsequent discriminators in our method.
Therefore, to investigate whether ${\cal S}_i$ contributes to
the improvement of both scores in all cases,
it is necessary to measure the change of the scores
for every state of ${\cal S}_j$ except ${\cal S}_i$.
Although strictly speaking,
measuring the effect on the strategy to sequentially activate the subsampling layers
does not fully illustrate the effectiveness of ${\cal S}_i$,
we can see whether all subsampling layers are effective,
at least in the general situation of activating them gradually.

We also confirmed the effectiveness of each subsampling layer for changes
in hyperparameters by setting $s_t$ to 2 and 4.
The batch size for these experiments was set to 16.
From the definition of $b_t$, the number of frames of four generated videos
in the case of $s_t=4$ is $[16,4,1,1]$, that is, enabling both ${\cal S}_1$
and ${\cal S}_2$ is the same as enabling all subsampling layers.
To see the effectiveness of ${\cal S}_3$ in $s_t=4$, instead,
we observed the change of IS and FID when ${\cal S}_3$ was activated
in the situation where ${\cal S}_1$ was enabled.

\begin{table}
\centering
{\renewcommand{\arraystretch}{1.2}
\begin{tabular}{c|ccc|ll}
Method & ${\cal S}_1$ & ${\cal S}_2$ & ${\cal S}_3$ & IS & FID \\ \hline \hline
Naive impl. & & & & $12.79 \pm .10$ & $7822 \pm 27$ \\
& \checkmark & & & $19.99 \pm .33$ & $4502 \pm 37$ \\
& \checkmark & \checkmark & & $21.38 \pm .38$ & $4192 \pm 45$ \\
Full & \checkmark & \checkmark & \checkmark & $22.81 \pm .52$ & $3639 \pm 15$  \\ \hline
\end{tabular}
}
\caption{Change in Inception Score and Fr\'echet Inception Distance when each subsampling layer is gradually enabled (batchsize=16, $s_t=2$).}
\label{table:is_change_2}
\end{table}

The result of $s_t=2$ is shown in \Table{is_change_2}.
By gradually activating ${\cal S}_i$, both IS and FID significantly improved.
We considered this is because that each sub-discriminator plays different roles
in improving the quality by enabling the subsampling layers, and IS and FID gradually increased.
Regarding the change in scores at each step, it can be seen that
IS and FID improved most when enabling ${\cal S}_1$.
It means that ${\cal S}_1$ contributed most to the improvement of the quality of videos in our method.

\begin{table}
\centering
{\renewcommand{\arraystretch}{1.2}
\begin{tabular}{c|ccc|ll}
Method & ${\cal S}_1$ & ${\cal S}_2$ & ${\cal S}_3$ & IS & FID \\ \hline \hline
Naive impl. & & & & $12.79 \pm .10$ & $7822 \pm 27$ \\
& \checkmark & & & $20.95 \pm .12$ & $4433 \pm 26$ \\
& \checkmark & & \checkmark & $19.37 \pm .43$ & $4628 \pm 20$ \\
Full & \checkmark & \checkmark &  & $22.69 \pm .47$ & $3715 \pm 19$ \\ \hline
\end{tabular}
}
\caption{Change in Inception Score and Fr\'echet Inception Distance when each subsampling layer is gradually enabled (batchsize=16, $s_t=4$).}
\label{table:is_change_4}
\end{table}

We also show the result of $s_t=4$ in \Table{is_change_4}.
As with the result of $s_t=2$,
IS and FID increased significantly by enabling ${\cal S}_i$ gradually.
In particular, the improvement of IS and FID when enabling ${\cal S}_1$
represents the importance of ${\cal S}_1$ in our method.
IS and FID did not improve when the number of frames of four generated videos
was changed from $[16,4,4,4]$ to $[16,4,1,1]$,
whereas both scores improved significantly
when changing from $[16,4,4,4]$ to $[16,4,1,1]$.
This implies that enabling ${\cal S}_2$ is more important than ${\cal S}_3$ for $s_t=4$.

\subsection{Frame subsampling in baseline models}

Although we show that the subsampling layer in our model has the effect of improving the quality of videos,
it is still unclear whether the subsampling layer is also effective in the conventional methods,
that is, the essential contribution of our method may be only in the subsampling layer.
To clarify that the contribution of this study is the combination of multi-scale model
and subsampling layers, we measured the change of IS and FID
when the subsampling layer was applied to the above two baseline models.

We describe the detail.
Unlike our proposed model that outputs multiple videos for training,
the generator in the baseline model generates only a single video.
Thus, from the network architecture of the baseline model,
it is clear that enabling ${\cal S}_2$ or ${\cal S}_3$ is equivalent to enabling only ${\cal S}_1$.
In this experiment, we focused on ${\cal S}_1$ and measured only the change of both scores
when ${\cal S}_1$ was activated.
As in the previous experiment, we used the two hyperparameters ($s_t=2, 4$)
for the subsampling layer. The batch size was 16.

\begin{table}[t]
\centering
{\renewcommand{\arraystretch}{1.2}
\begin{tabular}{c|ccc}
Method & Naive impl. & $s_t = 2$ & $s_t = 4$ \\ \hline \hline
3D dis. only & $11.10 \pm .16$ & $4.30 \pm .06$ & $1.73 \pm .01$ \\
3D + 2D dis. & $10.47 \pm .12$ & $4.97 \pm .04$ & $3.78 \pm .02$ \\ \hline
\end{tabular}
}
\caption{Changes in the Inception Score when applying a single frame subsampling layer to two baseline models. ``Naive impl.'' means a model without any subsampling layers.}
\label{table:frame_is}
\end{table}
\begin{table}[t]
\centering
{\renewcommand{\arraystretch}{1.2}
\begin{tabular}{c|ccc}
Method & Naive impl. & $s_t = 2$ & $s_t = 4$ \\ \hline \hline
3D dis. only & $8358 \pm 81$ & $11764 \pm 54$ & $13938 \pm 26$ \\
3D + 2D dis. & $8304 \pm 70$ & $11301 \pm 26$ & $11386 \pm 12$ \\ \hline
\end{tabular}
}
\caption{Changes in the Fr\'echet Inception Distance when applying a single frame subsampling layer to two baseline models.}
\label{table:frame_fid}
\end{table}

The results are shown in Tables \ref{table:frame_is} and \ref{table:frame_fid}.
Both IS and FID when activating the subsampling layer are
significantly worse than those of the naive implementation.
In particular, when the subsampling layer of $s_t=4$ was applied,
we observed that the training did not work at all for both models.
While the introduction of the subsampling layer saves computational cost and memory consumption,
the discriminator can identify the videos with reduced frame rates more easily than the original ones.
We considered that it led to a decrease in the IS and an increase in FID.
This result is in contrast to the result of our model,
which improved both computational cost and quality by enabling the subsampling layer,
and illustrates that our contribution is a combination of a multiscale model and subsampling layers.

\section{Discussion}

In this section, we summarize the empirical findings obtained through a series of experiments.

What contributed most to the improvement of the quality
is the combination of the subsampling layer and the multiscale model
(the quality cannot be dramatically improved by either the multiscale model or the subsampling layer).
The score can be improved by simply increasing the number of GPUs without the subsampling layers,
but its improvement is small.
The score improved significantly by enabling the subsampling layers.
However, this significant improvement does not occur in other models,
but only in our multiscale model.
That is, in order to improve the score in the task of video generation,
it is important not to introduce either one but to combine both.

The introduction of the subsampling layer leads not only to the improvement of the quality
but also to significantly saving the computational cost and memory consumption of the GPU
(c.f., \Table{performance}).
Even in situations where only a single GPU can be used due to budget constraints,
our method can efficiently solve high-resolution video generation problems
that could not be solved with naive models,
and its inception score is significantly higher than those of the existing methods.
We have confirmed that this dramatic increase in scores
cannot be achieved by only increasing the resolution with two baseline models.

Similar to the findings of BigGAN \cite{Brock2018}, the quality of the video can also be improved by increasing the batch size in our model.
We confirmed that an increase in $s_t$ leads to an improvement in the quality regardless of the batch size.
However, on the other hand, the generation of the first frames may be unstable depending on the dataset.
Since no such instability was observed in the case of $s_t=2$,
we consider that it would be appropriate to train the model with $s_t=2$ at an early stage,
gradually increase the value, and observe the change of IS and FID.

In our proposed method using multiple subsampling layers,
enabling the first subsampling layer (${\cal S}_1$) contributes the most to improving the score.
However, in our experiments, we did not see any decrease in scores even when the subsequent subsampling layers were enabled.
As there is no increase in computational cost and memory consumption by enabling the subsampling layer,
we consider that it would be better to enable all subsampling layers first,
and then to fine-tune by gradually activating the subsampling layers
if there are enough computing resources and time.

\subsection{Trials and their results}
We share a process of trial and error that led up to the development of our method.

\subsubsection{CLSTM layers}
We tried stacked Convolutional LSTM layers but observed that a single CLSTM with many channels worked better.
We also tried a dilated CLSTM layer, but the plain CLSTM was better. As with the above, we observed that simply increasing the number of channels in the single CLSTM contributed the most improvement in the score.

\subsubsection{Generation of longer videos}
We observed that our model sometimes succeeded to stably generate longer videos than the number of frames used for training (i.e., 16 frames) but sometimes generated broken videos.
Although this behavior depends on the hyperparameters and the dataset, it is unclear under what conditions our model can generate stably\footnote{Note that we mention videos after 16 frames. Up to 16 frames, our model generates stable videos regardless of the dataset.}.

\subsubsection{Spectral normalization}
We tried to insert the Spectral Normalization \cite{Miyato2018} into the 3D convolutional layers in the discriminator, but its performance decreased significantly.

\subsubsection{Gradient penalty}
Although we initially used a gradient penalty based on WGAN-GP \cite{Gulrajani2017} as a regularizer, but for simplicity we later adopted a simpler zero-centered gradient penalty.
Through the grid search, we finally confirmed that $\lambda = 0.5$ is the best but also observed that the training itself could be performed normally without the gradient penalty.
It may indicate that the method using multiple discriminators contributed to the stabilization of training.

\subsubsection{Batch normalization}
Mescheder et al. \cite{Mescheder2018}, the authors of the zero-centered gradient penalty, reported that they succeeded in generating high fidelity images without Batch Normalization layers in the generator, but in our experiments without Batch Normalization layers, the performance significantly decreased.

\subsubsection{3D discriminator}
The inception score when using a simple discriminator consisting of several 3D convolutional layers without any residual blocks is lower than the 3D ResNet discriminator, but the computational speed is faster. We used this simple discriminator temporarily for making the trial-and-error loops for searching a good model faster and adopted the 3D ResNet for the final evaluation.

\section{Conclusion}
In this paper we introduced the method of efficiently training the high-resolution model for video generation with GAN.
The main idea is to generate videos in different ways during training and inference.
During the training, the generator outputs multiple ``sparse'' samples
useful for the training at low computational cost
instead of directly generating high-resolution video.
For inference, the generator outputs high-resolution ``dense'' videos over time.
Using this method we can not only train our multi-scale model for video generation
while saving computational cost and memory consumption,
but the quality of our trained model is significantly superior to
the baseline models and the existing ones.
In both qualitative and quantitative experiments, we confirmed that
our method could output high-resolution videos with higher quality the conventional ones,
and the subsampling layers actually contribute to the improvement of the quality.

Our core idea may may not only to other fields that exploit videos
such as video prediction but also to other domains with time series
including audio generation.
We are planning to find more practical applications by exploring these possibilities.

\begin{acknowledgements}
We would like to acknowledge Takeru Miyato and Shoichiro Yamaguchi for helpful discussions.
We would like to acknowledge Daichi Suzuo for providing a tool to calculate the cost of computation and the amount of memory consumed.
We also would like to thank the developers of Chainer \cite{Tokui2015,Akiba2017}.
\end{acknowledgements}

\bibliographystyle{spmpsci}      

\bibliography{library}   

\end{document}